\pdfoutput=1

\documentclass[11pt]{article}

\usepackage{acl}

\usepackage{times}
\usepackage{latexsym}

\usepackage[T1]{fontenc}

\usepackage[utf8]{inputenc}

\usepackage{microtype}

\usepackage{inconsolata}

\usepackage{graphicx}

\usepackage{amsmath}
\usepackage{amssymb}
\usepackage{booktabs}
\usepackage{multicol}
\usepackage{multirow}
\usepackage{subcaption}

\title{Large Language Models Might Not Care What You Are Saying:\\
Prompt Format Beats Descriptions}

\author{Chenming Tang$^\dag$ \quad
Zhixiang Wang$^\dag$ \quad
Hao Sun \quad
Yunfang Wu\thanks{\ Corresponding author.} \\
  National Key Laboratory for Multimedia Information Processing, Peking University \\
  MOE Key Laboratory of Computational Linguistics, Peking University\\
  School of Computer Science, Peking University \\
  \texttt{\{tangchenming, ekko\}@stu.pku.edu.cn} \quad
  \texttt{\{2301213218, wuyf\}@pku.edu.cn}
}

\begin{document}
\maketitle

\def\thefootnote{$\dag$}\footnotetext{\ These authors contributed equally.}\def\thefootnote{\arabic{footnote}}

\begin{abstract}
With the help of in-context learning (ICL), large language models (LLMs) have achieved impressive performance across various tasks. However, the function of descriptive instructions during ICL remains under-explored. In this work, we propose an ensemble prompt framework to describe the selection criteria of multiple in-context examples, and preliminary experiments on machine translation (MT) across six translation directions confirm that this framework boosts ICL performance. But to our surprise, LLMs might not care what the descriptions actually say, and the performance gain is primarily caused by the ensemble format, since it could lead to improvement even with random descriptive nouns. We further apply this new ensemble framework on a range of commonsense, math, logical reasoning and hallucination tasks with three LLMs and achieve promising results, suggesting again that designing a proper prompt format would be much more effective and efficient than paying effort into specific descriptions.
\end{abstract}

\section{Introduction}
\label{sec:intro}

In-context learning (ICL) boosts the performance of large language models (LLMs) across numerous natural language processing (NLP) tasks, where LLMs are presented with in-context examples containing input and ground truth output \cite{NEURIPS2020_1457c0d6, dong2023survey}. Many works have verified the vital role of in-context examples in ICL \cite{wang-etal-2023-label, wei2023largerlanguagemodelsincontext}. However, \citeauthor{min-etal-2022-rethinking} \shortcite{min-etal-2022-rethinking} find that ground truth labels might not be the key to ICL performance on classification tasks.

The selection of in-context examples has been proven significant to the performance of ICL \cite{rubin-etal-2022-learning} and there have been various works on in-context example selection \cite{agrawal-etal-2023-context, li-etal-2023-unified, pmlr-v202-ye23c}. Besides diverse approaches of selecting examples, no existing work has tried to explicitly tell LLMs \emph{in what way those specific examples are selected}. We hypothesize that if LLMs are prompted with instructions describing the properties of selected in-context examples, they might learn better from these examples, since instruction following is one of LLMs' most important qualities nowadays \cite{NEURIPS2022_b1efde53, peng2023instructiontuninggpt4, zhang2024instructiontuninglargelanguage}. \citet{tang2024scoisyntaxaugmentedcoveragebasedincontext} prompt LLMs with examples selected based on both word-level and syntax-level criteria for machine translation (MT) for better ICL performance. This inspires us to tell LLMs where different in-context examples come from when they are selected by multiple methods. 

\begin{figure}[tbp]
    \centering
    \includegraphics[width=1\linewidth]{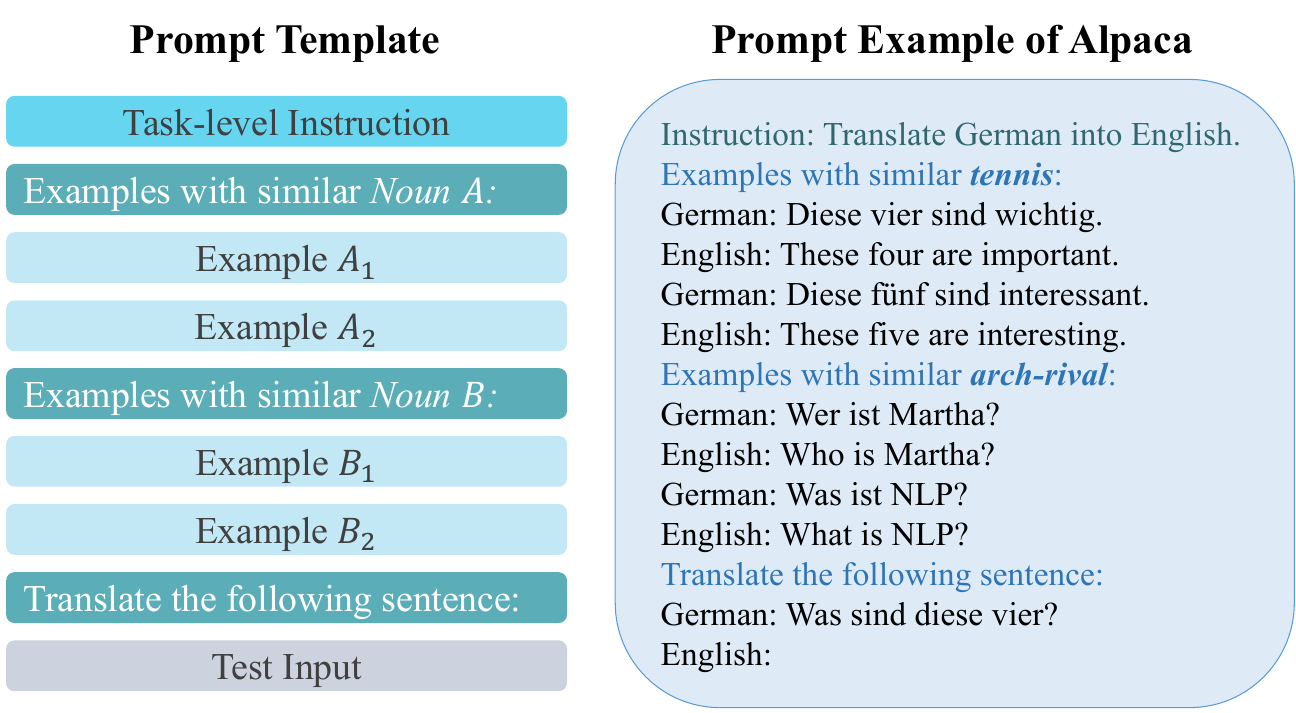}
    \caption{Template and Alpaca's example of \textit{Ensemble}.}
    \label{fig:ensemble-prompt}
\end{figure}

In our experiments on MT, we first select in-context examples based on lexical and syntactic similarity for each test input separately. Then we combine both to construct the complete set of examples, with half word-level examples and half syntax-level examples. Further, we devise a novel ensemble prompt framework (as shown in the left part "Prompt Template" of Figure \ref{fig:ensemble-prompt}), adding example-level instructions to describe that the following examples are with similar words or similar syntax. 

Experimental results on MT demonstrate that adding such ensemble prompt framework does improve LLMs' performance over conventional prompts. Meanwhile, we find that when the example-level descriptions do not correspond to the source of in-context examples or are completely nonsense, LLMs still benefit from the prompt. These surprising results indicate that in fact LLMs might not care what the descriptions say and are more sensitive to the prompt format. In other words, a proper format can be much more effective than well-designed descriptions in ICL.

To further verify the superiority of the ensemble framework, we present empirical evaluations on commonsense, math, logical reasoning and hallucination benchmarks (including nine datasets in total) across three small-scale LLMs (Alpaca, Llama3 and Mistral) and one large-scale LLM (GPT-3.5). The novel prompt framework is able to achieve promising results even with the descriptive nouns in the prompt being random nouns, further suggesting that a proper prompt format would be much more effective and efficient compared with laborious design of detailed and specific descriptions.

There are a few studies very related to our work. \citet{min-etal-2022-rethinking} find that the labels of in-context examples do not need to be correct for classification tasks. \citet{wei2023largerlanguagemodelsincontext} find that larger language models learn from in-context examples even when the labels are flipped or unrelated. \citet{nice} demonstrate that optimizing examples is less effective in some tasks given a high-quality task instruction. Our work is different from the above in that we focus on the meaning of \textbf{descriptions} rather than \textbf{labels} or \textbf{examples} in ICL and our finding is that the format of prompts is more important than carefully designed descriptions.

Our contributions can be summarized as follows:

\begin{itemize}
\item For the first time, we specifically analyze the effect of prompt descriptions on ICL performance and find that LLMs might not care what users actually say in descriptions, while they are more sensitive to the prompt format.
\item We present a simple yet effective prompt framework that is proven feasible on MT through comprehensive experiments across six translation directions. Promising experimental results on three LLMs further verify the superiority of the novel framework on a range of commonsense, math, logical reasoning and hallucination tasks.
\end{itemize}

Our code is available at \url{https://github.com/JamyDon/Format-Beats-Descriptions}.


\section{Prompting LLMs for MT}

Primarily, we focus on MT, a typical generation task. Recently, various approaches of selecting in-context examples have been proposed for MT \cite{agrawal-etal-2023-context, m-etal-2023-ctqscorer, tang2024scoisyntaxaugmentedcoveragebasedincontext}. However, no existing work has tried to make LLMs aware of \emph{in what way those specific in-context examples are selected}.

We assume that LLMs would perform better when they are told the reasons for selecting those examples. \citeauthor{tang2024scoisyntaxaugmentedcoveragebasedincontext} \shortcite{tang2024scoisyntaxaugmentedcoveragebasedincontext} select examples based on a combination of word-level and syntax-level criteria, which inspires us to present an ensemble prompt framework to make LLMs clearly know the reasons behind example selection.  In addition, to have a comprehensive idea of whether LLMs really know what is said in the descriptions, we design some prompt variants that are less meaningful or completely nonsense.

\subsection{In-context Example Selection for MT}
For word-level examples, we simply select them using BM25 \cite{Bassani_retriv_A_Python_2023}. For syntax-level examples, we use the top-$k$ polynomial algorithm proposed by \citeauthor{tang2024scoisyntaxaugmentedcoveragebasedincontext} \shortcite{tang2024scoisyntaxaugmentedcoveragebasedincontext} to convert dependency trees into polynomials and compute syntactic similarity based on the Manhattan distances \cite{Craw2017} between polynomial terms. For brevity, we denote the syntax-level algorithm by "Polynomial".

To combine word-level and syntax-level examples, we simply concatenate them. For example, the first and the remaining half of examples are selected by BM25 and Polynomial respectively.

\subsection{A New Ensemble Prompt Framework}

To maintain consistency, all our MT experiments use four in-context examples.

\begin{figure}[htbp]
    \centering
    \small
    \includegraphics[width=1\linewidth]{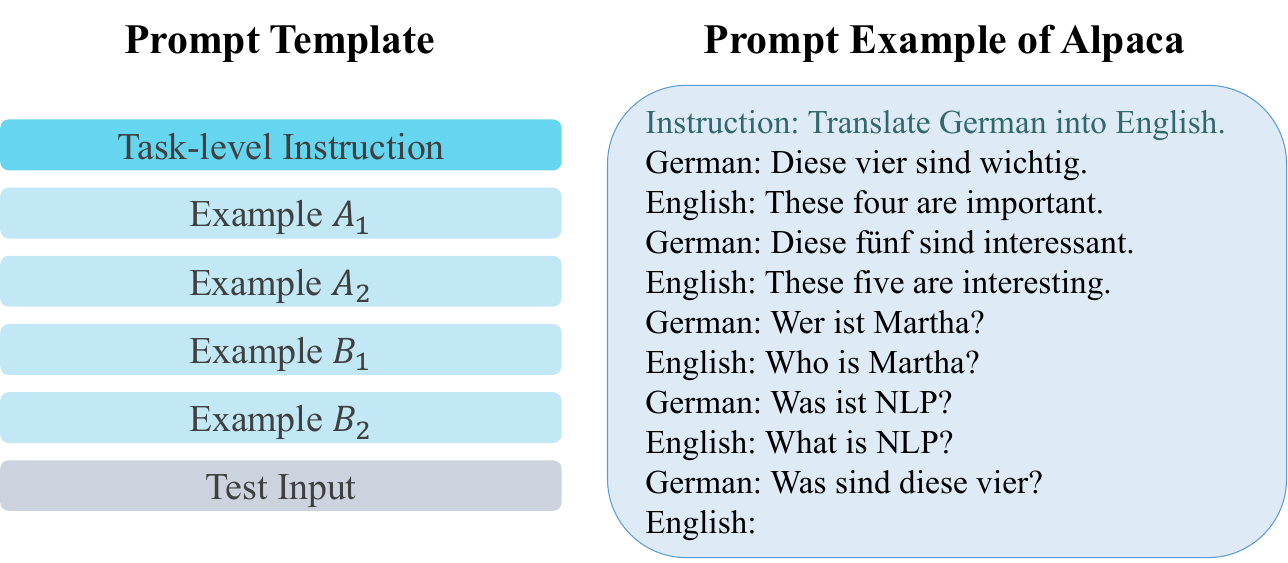}
    \caption{Template and Alpaca's example of \textit{Vanilla}.}
    \label{fig:vanilla-prompt}
\end{figure}

First of all, we use the most regular prompt without any example-level descriptions as baseline (referred to as \textit{Vanilla}), which is shown in Figure \ref{fig:vanilla-prompt}.

In the template, "Task-level Instruction" instructs the model to do the current task (MT here). "Example $A_i$" and "Example $B_i$" denote the $i$-th example from selection approach $A$ (\textit{e.g.}, BM25) and $B$ (\textit{e.g.}, Polynomial) respectively, all containing both source language inputs and target language translations. "Test Input" refers to the source language input of the test sample, which requires the LLM to translate it into the target language.

Then, we add example-level descriptions for examples from different selection approaches and explicitly instruct the LLM to translate the test input. This prompt framework is referred to as \textit{Ensemble} and is shown in Figure \ref{fig:ensemble-prompt} as presented in Section \ref{sec:intro}. "Noun $A$" and "Noun $B$" describe the examples from selection $A$ and $B$ respectively. For example, the two nouns can be "words" and "syntax" to properly describe examples selected by BM25 and Polynomial respectively. In this way, we can conveniently control the example-level descriptions to tell the LLM why those examples are used.

\subsection{Experimental Setup}
\label{subsec:setup}


\begin{table}[htbp]
\small 
  \centering
    \begin{tabular}{cccc}
    \toprule
    \textbf{Language} & \textbf{ISO Code} & \textbf{Dataset} & \textbf{\#Pairs (M)} \\
    \midrule
    German & DE    & Europarl & 1.8 \\
    French & FR    & Europarl & 1.9 \\
    Russian & RU    & ParaCrawl & 5.4 \\
    \bottomrule
    \end{tabular}
  \caption{Data statistics.}
  \label{tab:data}
\end{table}

\subsubsection{Datasets}
We perform evaluation on the \textit{devtest} set of FLORES-101 \cite{goyal-etal-2022-flores}, which contains 1012 sentences with translations in 101 languages. We experiment between English and three common languages: German, French and Russian. We use Europarl \cite{koehn-2005-europarl} for German and French and ParaCrawl \cite{banon-etal-2020-paracrawl} for Russian as example database, from which we select in-context examples. Detailed statistics are in Table \ref{tab:data}.

\subsubsection{Evaluation Metrics}
We report COMET \cite{rei-etal-2020-unbabels} scores from \texttt{wmt20-comet-da} \footnote{https://huggingface.co/Unbabel/wmt20-comet-da}, which is considered a superior metric for MT today \cite{kocmi-etal-2021-ship}.

\subsubsection{Language Models}
We experiment with two LLMs commonly used in MT: XGLM$_\text{7.5B}$ \cite{lin-etal-2022-shot} and Alpaca \cite{alpaca}. XGLM is a multilingual language model with 7.5B parameters supporting 30 languages that is frequently used in MT. Alpaca is a 7B LLM instruction-tuned from LLaMA \cite{touvron2023llamaopenefficientfoundation}.

\subsubsection{Example Selection}
To maintain consistency, all our MT experiments use four in-context examples (4-shot). We evaluate different ways of selecting examples for comparison. Note that if all four examples are selected by the same method, the first two are considered examples from $A$ and the last two are considered from $B$ in the \textit{Ensemble} template in Figure \ref{fig:ensemble-prompt}.

\textbf{Random:} The four examples are randomly sampled from the example database. We report the average result of three different random seeds.

\textbf{BM25:} We retrieve the top-$4$ matching examples for each test input using BM25 \cite{Bassani_retriv_A_Python_2023}.

\textbf{Polynomial:} It is rather time-consuming to retrieve examples from databases containing millions of data using the Polynomial algorithm. Following \citeauthor{tang2024scoisyntaxaugmentedcoveragebasedincontext} \shortcite{tang2024scoisyntaxaugmentedcoveragebasedincontext}, we instead re-rank the top-$100$ examples retrieved by BM25 using Polynomial and the top-$4$ are used as final in-context examples.

\textbf{BM25 + Polynomial:} To combine examples with both lexical and syntactic similarity, we simply concatenate examples from BM25 and Polynomial. Specifically, the first two examples are from BM25 and the remaining two are from Polynomial.

\textbf{Polynomial + BM25:} The first two examples are from Polynomial and the remaining two are from BM25.

\subsubsection{Prompts}
We design various prompts to explore whether LLMs can benefit from explicit descriptions of examples and whether they really understand the meaning of descriptions.

\textbf{\textit{Vanilla}:} The normal prompt without any example-level descriptions as shown in Figure \ref{fig:vanilla-prompt}.

\begin{figure}
    \centering
    \begin{subfigure}[t]{0.45\textwidth} 
        \centering
        \includegraphics[width=\linewidth]{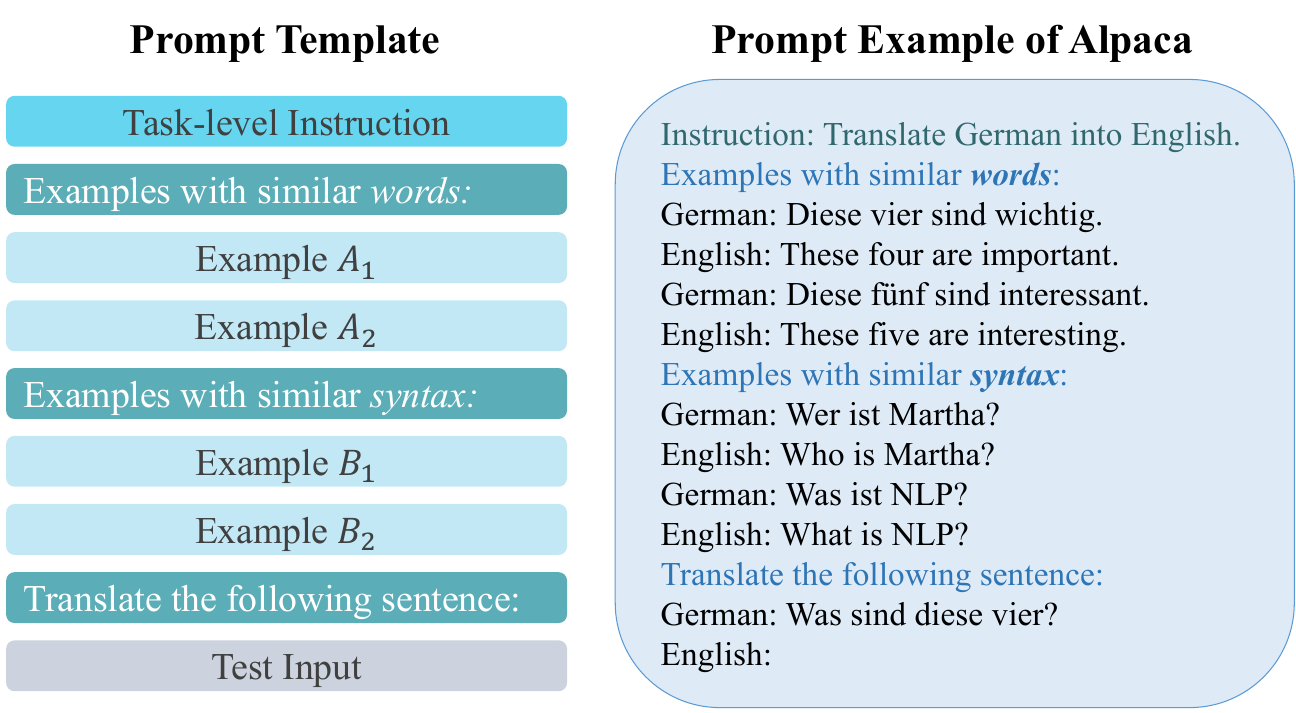} 
        \caption{\textit{Ensemble} (Word + Syntax).}
        \label{subfig:EWS}
    \end{subfigure}

    \begin{subfigure}[t]{0.45\textwidth} 
        \centering
        \includegraphics[width=\linewidth]{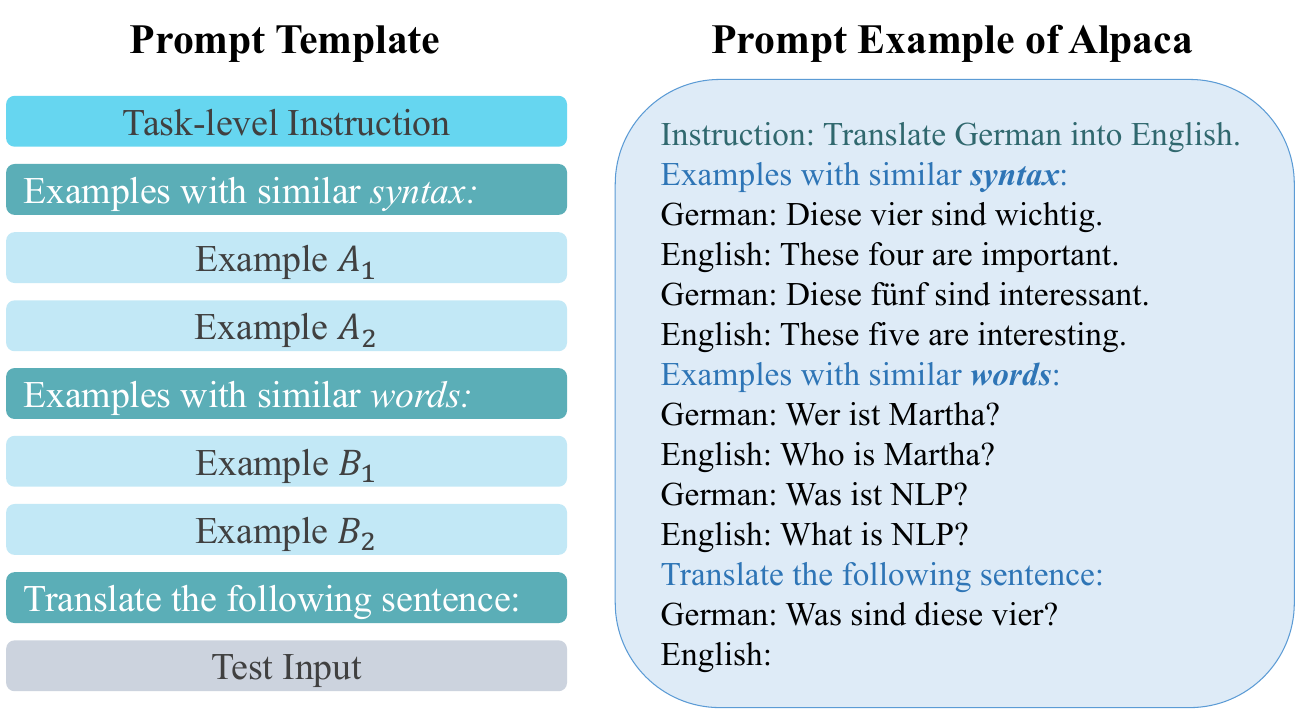} 
        \caption{\textit{Ensemble} (Syntax + Word).}
        \label{subfig:ESW}
    \end{subfigure}

    \begin{subfigure}[t]{0.45\textwidth} 
        \centering
        \includegraphics[width=\linewidth]{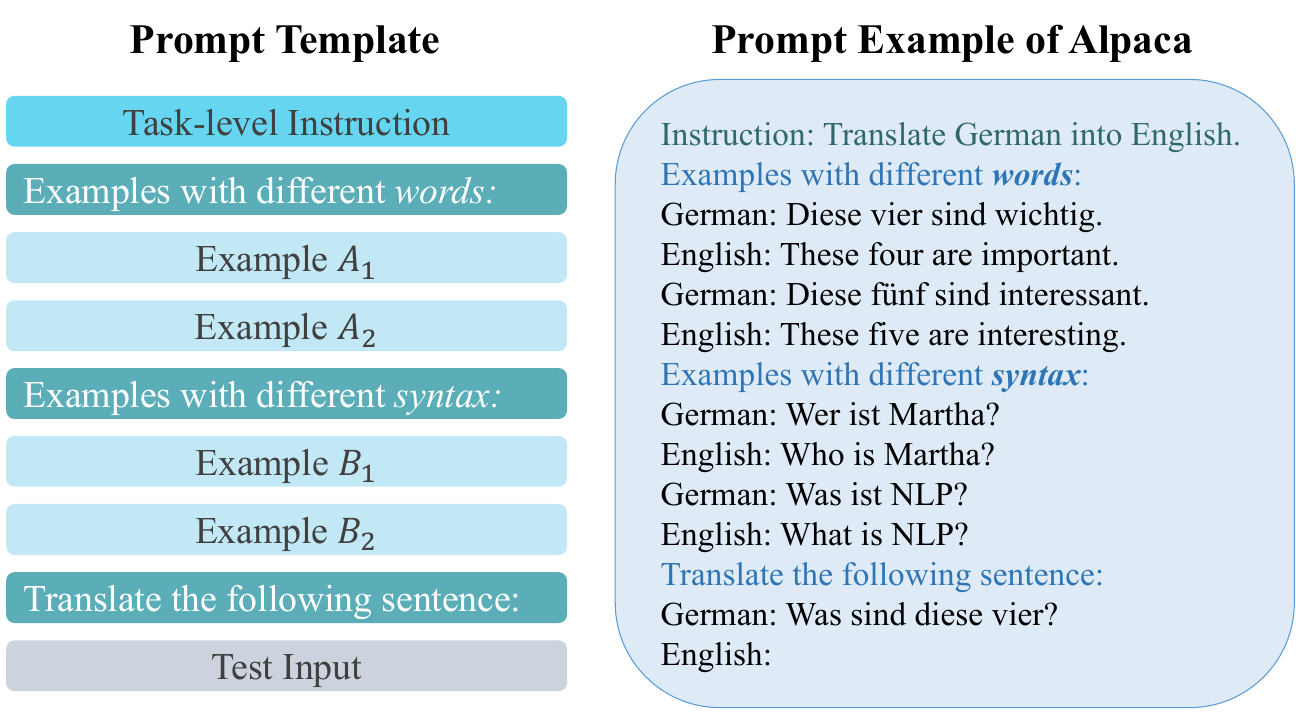} 
        \caption{Different \textit{Ensemble} (Word + Syntax).}
        \label{subfig:DiffEWS}
    \end{subfigure}

    \begin{subfigure}[t]{0.45\textwidth} 
        \centering
        \includegraphics[width=\linewidth]{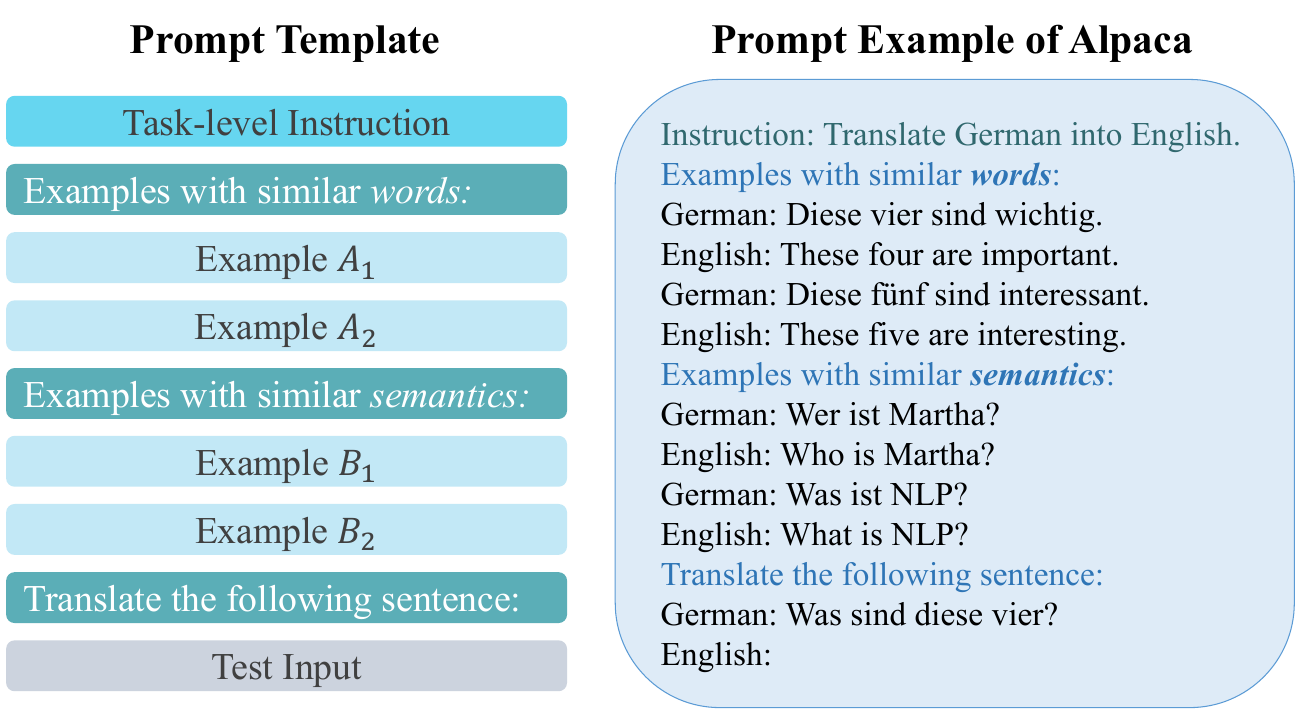} 
        \caption{\textit{Ensemble} (Word + Semantics).}
        \label{subfig:EWSem}
    \end{subfigure}

    \begin{subfigure}[t]{0.45\textwidth} 
        \centering
        \includegraphics[width=\linewidth]{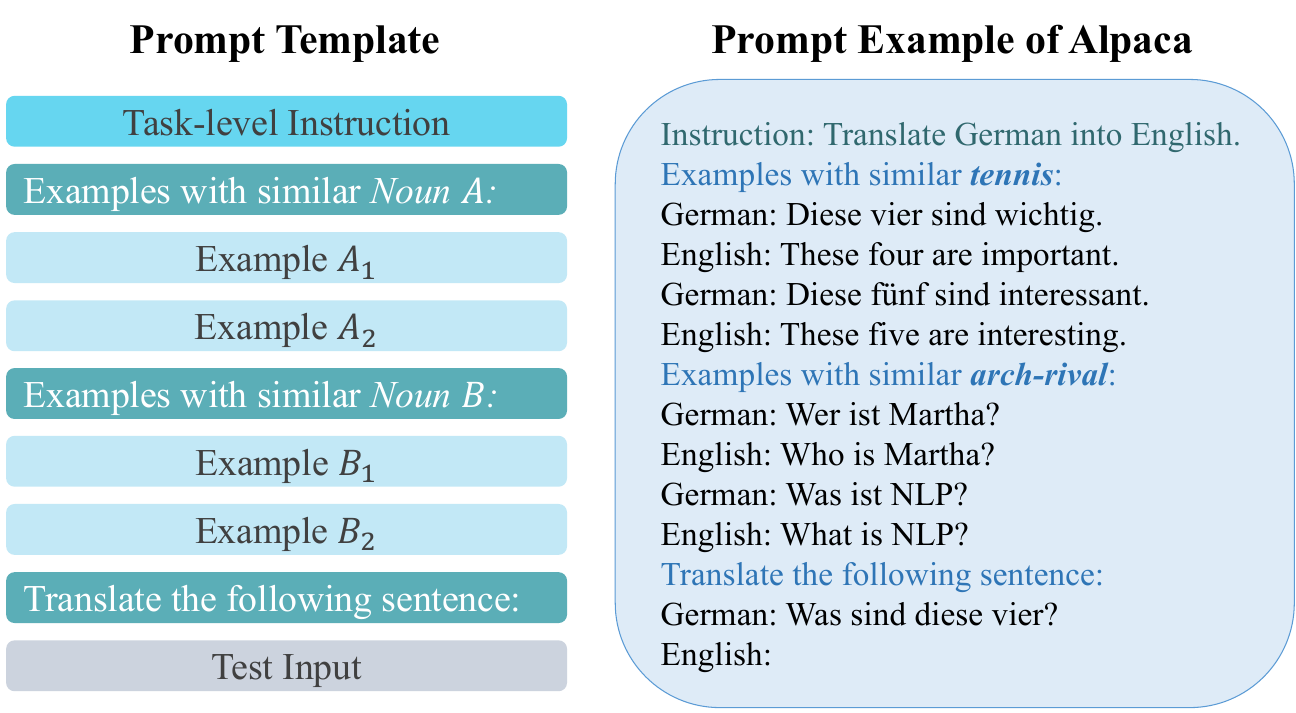} 
        \caption{\textit{Ensemble} (Random + Random).}
        \label{subfig:ERR}
    \end{subfigure}
    \caption{Templates and Alpaca's examples of \textit{Ensemble} prompts.}
    \label{fig:ens-template}
\end{figure}

\textbf{\textit{Ensemble} (Word + Syntax):} Shown in Figure \ref{subfig:EWS}, Noun $A$ and Noun $B$ are "words" and "syntax" respectively, which semantically corresponds to BM25 + Polynomial examples but is converse to Polynomial + BM25. 

\textbf{\textit{Ensemble} (Syntax + Word):} Shown in Figure \ref{subfig:ESW}, Noun $A$ and Noun $B$ are "syntax" and "words" respectively, which semantically matches Polynomial + BM25 examples but mismatches BM25 + Polynomial. 

\textbf{Different \textit{Ensemble} (Word + Syntax):} Shown in Figure \ref{subfig:DiffEWS}, Noun $A$ and Noun $B$ are still "words" and "syntax" respectively but the qualifier "similar" is replaced with "different". This aims to find out whether LLMs pay attention to the meaning of "different/similar" and care the semantics of descriptions.

\textbf{\textit{Ensemble} (Word + Semantics):} Shown in Figure \ref{subfig:EWSem}, Noun $A$ and Noun $B$ are "words" and "semantics" respectively, which does not semantically match any of our example selection methods.

\textbf{\textit{Ensemble} (Random + Random):} Shown in Figure \ref{subfig:ERR}, for each input, Noun $A$ and Noun $B$ are different random English nouns sampled using Wonderwords~\footnote{https://github.com/mrmaxguns/wonderwordsmodule}, aiming to explore LLMs' understanding of descriptions.


\begin{figure*}[htbp]
    \centering
    \includegraphics[width=1\linewidth]{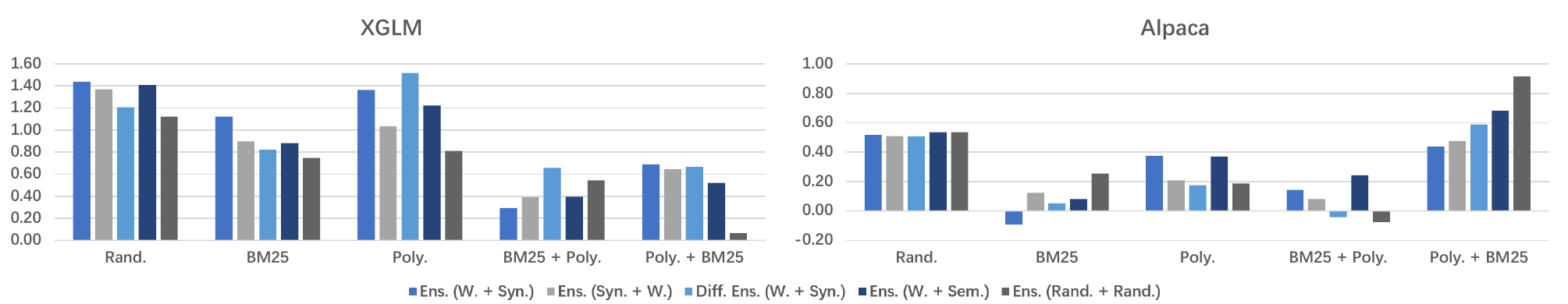}
    \caption{Main results on XGLM and Alpaca, showing the performance gain of different prompts over the \textit{Vanilla} prompt, averaged over all six translation directions. Each cluster presents the results of a selection of in-context examples and each bar in it presents the result of a prompt. "Ens.", "W.", "Syn.", "Sem.", "Diff.", "Rand.", "Poly." refer to "\textit{Ensemble}", "Word", "Syntax", "Semantics", "Different", "Random", "Polynomial", respectively.}
    \label{fig:mt-main}
    \vspace{-0.4cm}
\end{figure*}

\subsection{Main Results}
To give a quick view of LLMs' MT performance, Table \ref{tab:baseline} shows the COMET scores of \textit{Vanilla} baselines averaged over six translation directions. 

\begin{table}[htbp]
    \centering
    \small
    \begin{tabular}{lcc}
        \toprule
        \textbf{Example Selection} & \textbf{XGLM} & \textbf{Alpaca} \\
        \midrule
        Random & 54.07 & 55.42 \\
        BM25 & 55.00 & 56.27 \\
        Polynomial & 55.52 & 56.13 \\
        BM25 + Polynomial & 56.17 & 56.18 \\
        Polynomial + BM25 & 56.18 & 55.49 \\
        \bottomrule
    \end{tabular}
    \caption{Results of \textit{Vanilla} baselines of XGLM and Alpaca with different example selection methods, averaged over six translation directions.}
  \label{tab:baseline}
\end{table}

Main results are shown in Figure \ref{fig:mt-main}. For convenient comparison, we present the performance gain of different \textit{Ensemble} prompts over \textit{Vanilla} with different selections of in-context examples and the results are averaged over six translation directions. For detailed results of different translation directions, please refer to Appendix~\ref{sec:full}.

As can be seen from the results, those \textbf{"correct"} prompts, exactly corresponding to the selection of in-context examples (\textit{e.g.}, \textit{Ensemble} (Word + Syntax) with BM25 + Polynomial examples and \textit{Ensemble} (Syntax + Word) with Polynomial + BM25 examples), do bring some help as expected. However, when the prompt does not correspond to the selection of examples (\textit{i.e.}, is \textbf{"incorrect"}), the performance improves as well and sometimes even more than those "correct" cases. For example, on XGLM with BM25 + Polynomial examples, \textit{Ensemble} (Syntax + Word) improves more than \textit{Ensemble} (Word + Syntax), even though the former is completely reversed. On Alpaca with BM25 + Polynomial examples, \textit{Ensemble} (Word + Semantics) improves more than \textit{Ensemble} (Word + Syntax), albeit the examples with similar syntax do not necessarily bear similar semantics. More interestingly, Different \textit{Ensemble} (Word + Syntax), telling the LLM that the in-context examples are with different properties, is able to beat "correct" prompts sometimes (\textit{e.g.}, on XGLM with BM25 + Polynomial examples and Alpaca with Polynomial + BM25 examples).

Surprisingly, no matter how in-context examples are selected and whether the prompts are "correct", \textit{Ensemble} prompts bring improvement in most cases. Even \textit{Ensemble} (Random + Random), in which example-level descriptions are with random nouns and could be completely nonsense (like "examples with similar nobody"), brings improvement in most cases, especially obtaining the most gain on Alpaca with Polynomial + BM25 examples compared with other prompts, correct or incorrect. These results indicate that LLMs might not really take the example-level descriptions into consideration during ICL. In other words, they might not necessarily care what users say in the descriptions.

Compared with proper descriptions, it seems the format of prompts matters more. For example, on Alpaca with Random examples, no matter what the example-level descriptions say, all \textit{Ensemble} prompts bring nearly equal improvement over \textit{Vanilla}. This indicates that \textit{Ensemble} is a superior format compared with \textit{Vanilla} in this case.

To sum up, the experimental results on MT suggest that a proper prompt format leads to better ICL performance of LLMs while a careful design of descriptions might be less effective.

\subsection{Ablation Study}

To better understand how the \textit{Ensemble} format brings improvement, we perform ablation experiments over the organization of the prompt:

\textbf{\textit{Ensemble} (Random + Random):} The \textit{Ensemble} prompt with random nouns in the example-level descriptions as described in Section \ref{subsec:setup}.

\textbf{\textit{Single} (Random):} Organized based on Figure \ref{fig:ensemble-prompt}, but the second description is removed. There is only one example-level description above the four examples, where Noun $A$ is a random noun.

\textbf{\textit{Single} (Example):} Organized based on Figure \ref{fig:ensemble-prompt}, but the second description is removed. There is only one example-level description above the four examples, being "Examples:" only, without any further descriptions. This prompt only informs the LLM that the following four instances are examples and does not describe their properties.

\textbf{\textit{Vanilla} (Translate):} Organized based on Figure \ref{fig:ensemble-prompt}, but both the two descriptions are removed. The only difference with \textit{Vanilla} is the translation instruction "Translate the following sentence:" before the test input. This prompt only informs the LLM to translate the test input and tells nothing about the in-context examples.

Detailed templates and examples of the above prompts are presented in Appendix \ref{subapp:prompt-ablation}.

\begin{figure}[htbp]
    \centering
    \includegraphics[width=1\linewidth]{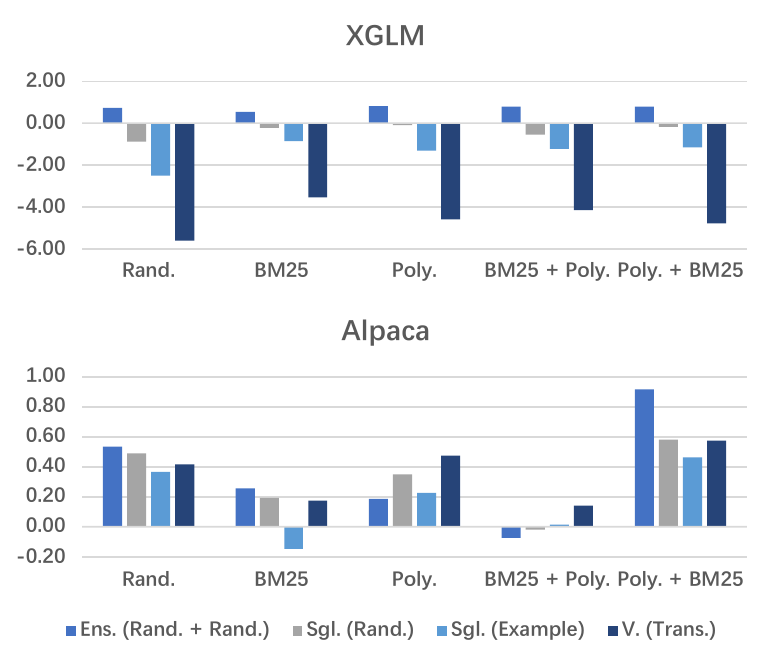}
    \caption{Ablation studies over the organization of the prompt, showing the performance gain of different prompts over \textit{Vanilla}, averaged over all six translation directions. "Rand.", "Poly.", "Ens.", "Sgl.", "V.", "Trans." refer to "Random", "Polynomial", "\textit{Ensemble}", "\textit{Single}", "\textit{Vanilla}", "Translate", respectively.}
    \label{fig:mt-ablation}
\end{figure}

\begin{figure*}[ht]
    \centering
    \includegraphics[width=0.95\linewidth]{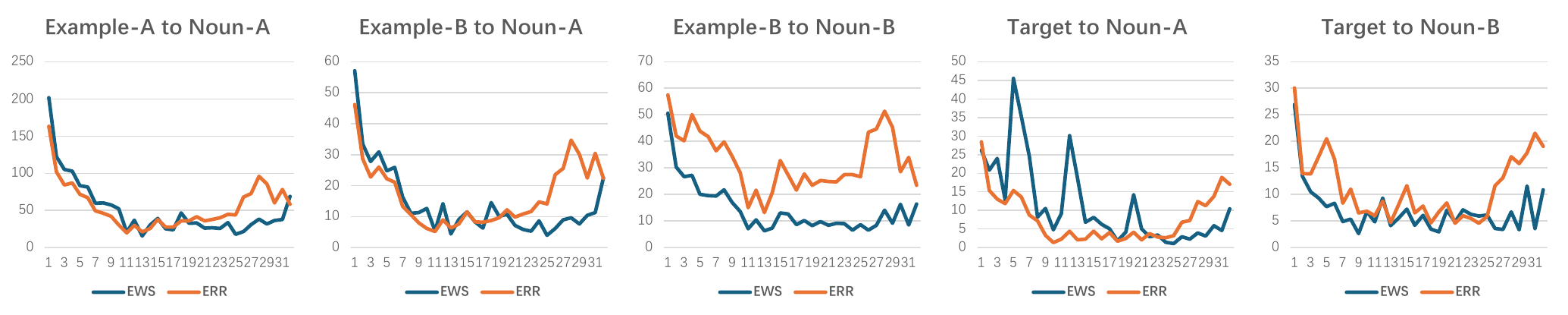}
    \caption{Attention weights (×1e-4) on XGLM of all 32 layers with BM25 + Polynomial examples. \textit{EWS} and \textit{ERR} denotes \textit{Ensemble} (Word + Syntax) and \textit{Ensemble} (Random + Random) respectively.}
    \label{fig:attn-xglm}
\end{figure*}

Results are presented in Figure \ref{fig:mt-ablation}, showing that removing one or two example-level descriptions or removing the random noun describing the property of in-context examples hurt the performance gain in most cases. On XGLM, only the original \textit{Ensemble} format performs better than \textit{Vanilla}. Alpaca exhibits an abnormal trend when prompted with Polynomial and BM25 + Polynomial examples, where \textit{Ensemble} (Random + Random) cannot outperform other prompts. This may be due to that Alpaca is instruction-tuned and the \textit{Single} or \textit{Vanilla} (Translate) prompts are also friendly to it in some cases because of the post-training stage. But overall, \textit{Single} (Random), \textit{Single} (Example) and \textit{Vanilla} (Translate) still bring less improvement than \textit{Ensemble} (Random + Random) in more than half of the cases.

Ablation experiments suggest that in MT, our proposed \textit{Ensemble} is a relatively superior prompt format, performing better than other variants.

\subsection{Analysis via Attention Weights}
\label{sec:attention}

To have a better idea of the internal mechanism of LLMs when prompted with different prompts, we compute the attention weights between different prompt components. We focus on three components: in-context examples (from $A$ or $B$, denoted by "Example-A" and "Example-B"), the target position (denoted by "Target") where the model starts to generate predictions (following \citeauthor{wang-etal-2023-label} \shortcite{wang-etal-2023-label}, we use the final token in the input) and the two descriptive nouns ("Noun-A" and "Noun-B"). We obtain the attention weights averaged over all attention heads from the attention matrix across all the layers.
All the results are averaged over all six language directions.

Results comparing \textit{Ensemble} (Word + Syntax) (\textit{EWS}) and \textit{Ensemble} (Random + Random) (\textit{ERR}) on XGLM with BM25 + Polynomial examples are presented in Figure \ref{fig:attn-xglm} (for results on Alpaca, refer to Appendix \ref{sec:attn-alpaca}). If the model really cares what the descriptions say, its attention to meaningful descriptive nouns (in \textit{EWS}) should be much greater than those meaningless (in \textit{ERR}). However, in most cases, \textit{EWS} performs no higher than \textit{ERR}, indicating that the model does not really care what the descriptive nouns actually are. "Target to Noun-A" is a special case, where \textit{EWS} is high in shallow layers. But in deeper layers, \textit{EWS} falls behind and \textit{ERR} takes the lead. This shows that the model might pay more attention to the meaningful noun when understanding the context in shallow layers but gradually forgets it when it comes to generation in deeper layers. In a word, the attention weights further confirm our claim that LLMs do not really care what the descriptive nouns are in most cases.

\subsection{Discussion}

Above results show that LLMs benefit from our \textit{Ensemble} prompts in most cases. However, the benefit comes from a proper format rather than the meaningful descriptions (\textit{e.g.}, "similar words" and "similar syntax"). This demonstrates that LLMs might not care what users say in the descriptions but is more sensitive to the format of prompts. In other words, designing a proper prompt format would be more efficient than 
paying a lot of effort into looking for a perfect description.

In the next section, we apply \textit{Ensemble} format to more tasks to further verify its generalizability.

\section{Generalizing the New Ensemble Prompt Framework to More Tasks}

To further verify our conclusion obtained from MT that our proposed \textit{Ensemble} framework improves ICL even when the example-level descriptions are incorrect or meaningless, we perform the comparison between \textit{Vanilla} and \textit{Ensemble} (Random + Random), which we would refer to as \textit{ERR}, on more types of tasks across different language models.

\begin{figure*}[htbp]
    \centering
    \includegraphics[width=0.95\linewidth]{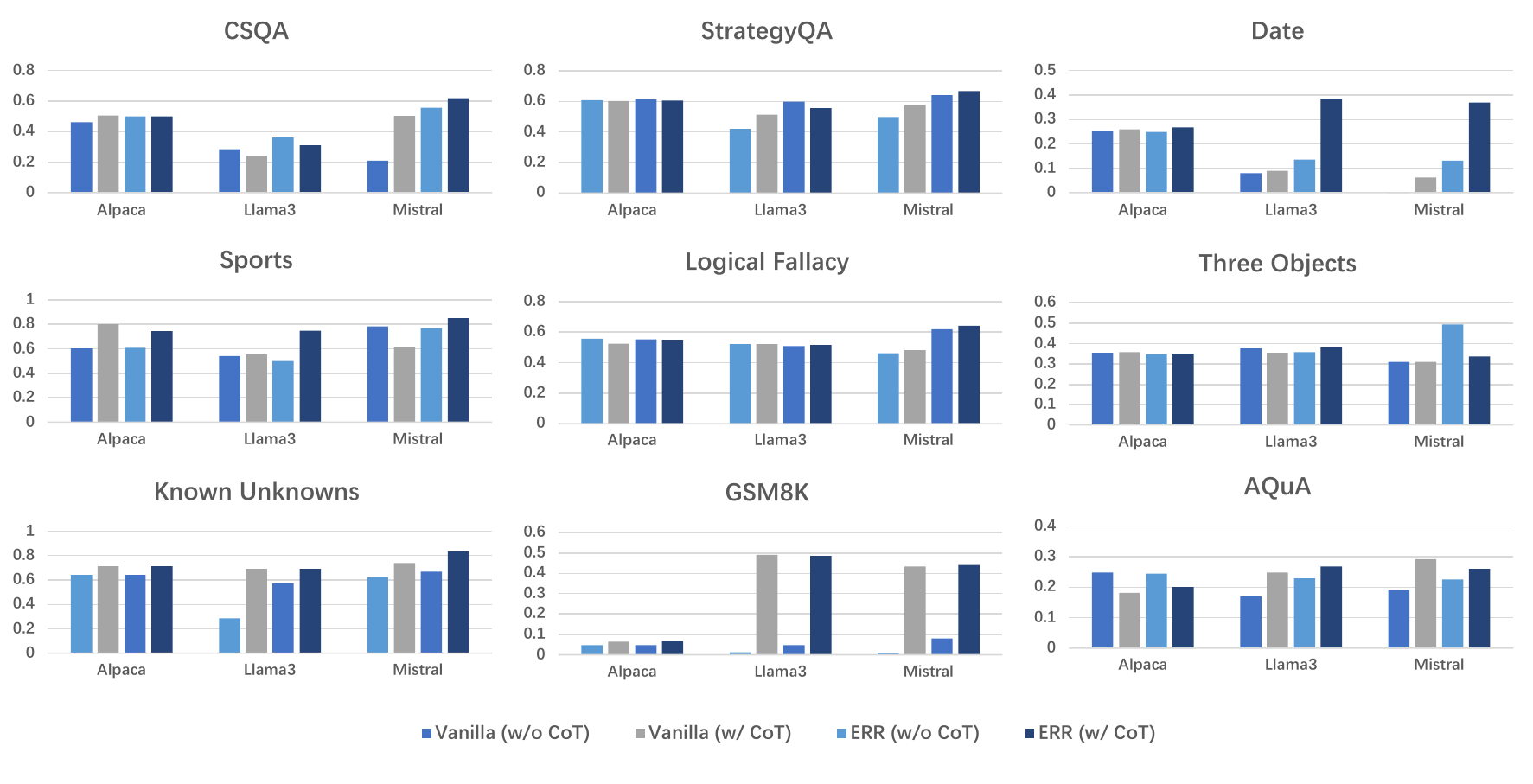}
    \caption{Results on nine datasets across three small-scale models. In the "Date" subplot, the score of Mistral under the \textit{Vanilla} prompt is too low to be a visible bar in the chart.}
    \label{fig:7b}
\end{figure*}

\subsection{Experimental Setup}

\subsubsection{Datasets}
We use a total of nine benchmarks, covering four task types: commonsense QA, logical reasoning, arithmetic reasoning, and hallucination detection. 

For commonsense QA, we adopt four datasets.
The widely-used CSQA \cite{csqa} features commonsense questions about the world involving complex semantics requiring prior knowledge. StrategyQA \cite{strategyqa} challenges models to infer implicit reasoning steps using a strategy to answer questions. We also choose two specialized evaluation sets from BIG-bench \cite{bigbench}: Date Understanding, which asks models to infer the date from a context, and Sports Understanding, which involves assessing the plausibility of sentences related to sports.

For logical reasoning task, we choose Logical Fallacy and Three Objects (a subset of Logical Deduction) from Big-bench \cite{bigbench}. Logical Fallacy aims to test the model's ability to identify whether there are fallacies in a given logical reasoning, and Three Objects requires the model to infer the order of a sequence of objects from a set of minimal conditions.

To explore the performance of \textit{ERR} on math word problems, we adopt the following two datasets: GSM8K \cite{gsm8k}, which consists of high quality free-response grade school math problems, and AQuA \cite{aqua}, containing the algebraic word problems in the form of multiple-choice questions.

In addition, to explore whether \textit{ERR} could alleviate LLMs' hallucination, we choose Known Unknowns from Big-bench \cite{bigbench}.

\begin{table}[htbp]
    \centering
    \small
    \begin{tabular}{lc}
        \toprule
        \textbf{Dataset} & \textbf{Test Inputs} \\
        \midrule
        CSQA & 1221 \\
        StrategyQA & 1012 \\
        Date & 365 \\
        Sports & 996 \\
        Logical Fallacy & 1012 \\
        Three Objects & 296 \\
        Known Unknowns & 42 \\
        GSM8K & 1319 \\
        AQuA & 254 \\
        \bottomrule
    \end{tabular}
    \caption{Number of test inputs for each dataset.}
  \label{tab:num_test_inputs}
\end{table}

The number of test inputs for each dataset is listed in Table \ref{tab:num_test_inputs}. Details of splitting training set (example database) and test set are in Appendix \ref{app:data}.

\subsubsection{Evaluation Metric}
These nine datasets are either in the form of multiple-choice questions or free-response questions with standard answers, so we use accuracy as the metric for all of them.

\subsubsection{Language Models}
We experiment with both instruction-tuned and non-instruction-tuned models to see whether our findings could extend to different kinds of models. We evaluate three frequently used open source LLMs with around 7B parameters, including Alpaca~\cite{alpaca}, Llama3~\cite{grattafiori2024llama3herdmodels}, and Mistral~\cite{mistral}, among which Llama3 is a base model before instruction tuning. To assess the effect of \textit{ERR} on more powerful models, we also evaluate GPT-3.5~\cite{NEURIPS2022_b1efde53}~\footnote{We choose this model because it is a commonly used cost-effective API-based LLM and a \textit{de facto} black box baseline.}. We use Llama-3.1-8B, Mistral-7B-Instruct-v0.2 and gpt-3.5-turbo-0125~\footnote{https://openai.com/api/} for Llama3, Mistral and GPT-3.5 respectively.

\subsubsection{Example Selection}
Note that randomly selected examples combined with \textit{ERR} have already brought non-trivial improvements to MT. Therefore, for each dataset discussed in this section, we randomly select a uniform set of examples (4-shot) for all test inputs without applying any carefully designed selection method, in order to focus on and verify the simple yet effective and universal nature of \textit{ERR}.

\subsubsection{Prompts}
We compare \textit{ERR} with \textit{Vanilla} across different datasets and LLMs. Given that these tasks usually involve reasoning, on which chain-of-thought (CoT) is commonly utilized \cite{cot}, we experiment both without CoT ("w/o CoT", which are identical to the original templates) and with CoT ("w/ CoT"). This allows us to examine both the orthogonality and compatibility with CoT of \textit{ERR} , as well as assess its performance across various models and tasks. Specifically, we evaluate \textit{Vanilla} (w/o CoT), \textit{Vanilla} (w/ CoT), \textit{ERR} (w/o CoT), and \textit{ERR} (w/ CoT). Due to space constraints, examples of prompt templates discussed in this section are provided in Appendix \ref{app:prompt-reasoning}.

\begin{figure}[htbp]
    \centering
    \includegraphics[width=1\linewidth]{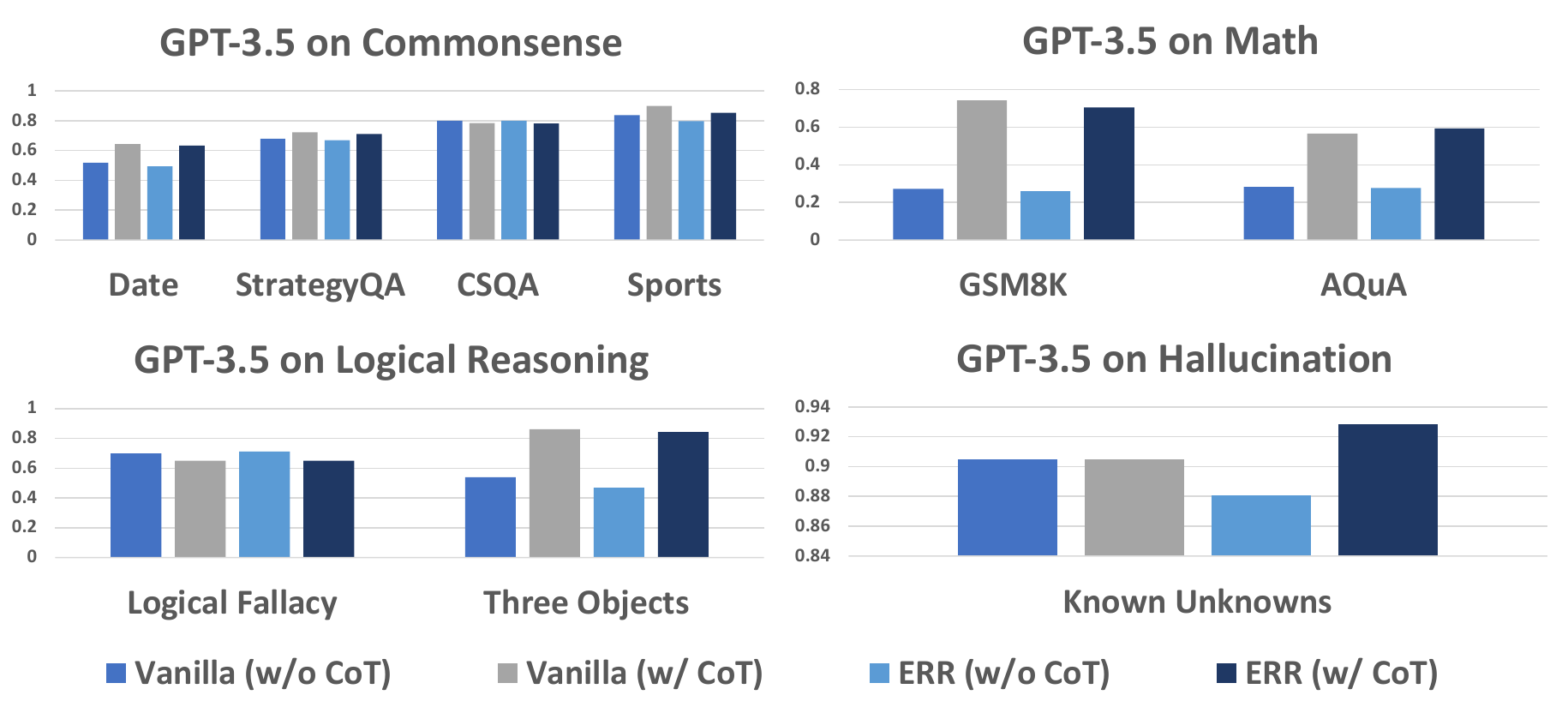}
    \caption{Results of the four types of tasks on GPT-3.5.}
    \label{fig:gpt}
\end{figure}

\subsection{Results of Small-scale Models}
\label{sec:small-reasoning}
Results across all nine datasets and three small-scale models (Alpaca, Llama3 and Mistral) are illustrated in Figure \ref{fig:7b}. Detailed results are presented in Appendix~\ref{sec:full}.

The results demonstrate that \textit{ERR} (w/ CoT), achieved by integrating CoT with our proposed prompt framework, either significantly outperforms or matches \textit{Vanilla} (w/ CoT) in 25 out of 27 experiments (covering nine datasets and three models). The exceptions are Alpaca on the Sports dataset and Mistral on the AQuA dataset, where \textit{ERR} (w/ CoT) shows somewhat lower performance compared to \textit{Vanilla} (w/ CoT). When CoT is not incorporated, \textit{ERR} generally performs much better than or on par with \textit{Vanilla}, except for the Sports dataset with Llama3, where \textit{ERR} performs a little poorer.

Surprisingly, \textit{ERR} (w/o CoT) sometimes even surpasses \textit{Vanilla} (w/ CoT), suggesting that the \textit{ERR} framework alone can offer more improvements than CoT. This highlights the value of \textit{ERR} and reaffirms that the format plays a crucial role in enhancing LLMs' problem-solving capabilities. In terms of models, the performance of \textit{ERR} on Alpaca is far less impressive than on Llama3 and Mistral, which may be because Alpaca has strong instruction-following capabilities and is more robust to different prompts.

In summary, without using any carefully designed selection methods, directly filling the randomly selected examples into the \textit{ERR} framework brings significant improvement to various reasoning tasks and even alleviates the hallucination of models in most cases, no matter how meaningless and incorrect the example-level descriptions are. Moreover, \textit{ERR} can work perfectly with CoT. Therefore, at least for relatively small models, this simple yet effective trick is worth introducing into prompt engineering for various tasks.

We also experiment with Llama2 \cite{llama2} and the results are in Appendix \ref{app:llama2}. The overall trend is consistent with Llama3.
\subsection{Results of GPT-3.5}

As shown in Figure \ref{fig:gpt}, \textit{ERR} performs similarly to \textit{Vanilla} across every dataset using GPT-3.5. Although the \textit{ERR} format does not bring significant improvement to these tasks with GPT-3.5 and Alpaca (as shown in Figure \ref{fig:7b}), the fact remains that the incorrect or meaningless example-level descriptions caused by random nouns do not have much negative impact on GPT-3.5, a sufficiently powerful model, or Alpaca, which has strong instruction-following capabilities. In some cases, it even slightly improves performance (\textit{e.g.}, \textit{ERR} (w/ CoT) outperforms \textit{Vanilla} (w/ CoT) on AQuA and Known Unknowns). In other words, LLMs might not care what users actually say to describe the provided examples while they are more sensitive to the format of prompts, which is in line with our findings obtained from MT.

\subsection{Discussion}
Based on the experiments conducted on both small-scale and large-scale models, we can conclude that \textit{ERR} is a simple yet practical and universal prompt framework. It can enhance problem-solving capabilities of small models and be applied to large models without the risk of performance degradation due to the meaningless noise within it. In other words, there might be less need to meticulously select examples or design detailed descriptions. Instead, you can uniformly and efficiently apply \textit{ERR} to various tasks with different models.

As analyzed in Section~\ref{sec:attention}, the \textit{ERR} framework can work because LLMs pay less attention to the descriptive nouns while being more sensitive to the overall prompt format. We conjecture that the underlying reason could be that LLMs have been presented with many patterns similar to \textit{ERR} during pre-training and thus perform better when presented with \textit{ERR} prompts~\cite{parallel}. However, due to lack of access to the pre-training process of LLMs (either open-source or close-source), we cannot further validate our conjecture more solidly and our understanding of the deeper mechanism remains limited to superficial analysis, which is one of the limitations of this work.

\section{Related Work}

\paragraph{In-context Example Selection}

\citeauthor{rubin-etal-2022-learning} \shortcite{rubin-etal-2022-learning} suggest that LLMs' ICL performance strongly depends on the selection of in-context examples. In consequence, many works have been trying to explore ways of selecting better in-context examples in recent years. \citeauthor{li-etal-2023-unified} \shortcite{li-etal-2023-unified} train a unified in-context example retriever across a wide range of tasks. \citeauthor{pmlr-v202-ye23c} \shortcite{pmlr-v202-ye23c} select examples based on both relevance and diversity, with the help of determinantal point processes. \citeauthor{agrawal-etal-2023-context} \shortcite{agrawal-etal-2023-context} ensure n-gram coverage to select better examples for MT. \citeauthor{m-etal-2023-ctqscorer} \shortcite{m-etal-2023-ctqscorer} train an in-context example scorer for MT based on several features. \citeauthor{tang2024scoisyntaxaugmentedcoveragebasedincontext} \shortcite{tang2024scoisyntaxaugmentedcoveragebasedincontext} combine both word-level and syntax-level coverage when selecting examples for MT.

\paragraph{Mechanism of In-context Learning}

With the popularity of ICL, there have been numerous studies on analyzing the mechanism of ICL. One stream of these studies focuses on explaining the essence of ICL, relating ICL to gradient descent \cite{von2023transformers}, implicit Bayesian inference \cite{xie2022an}, induction heads completing token sequences based on similar context \cite{olsson2022incontextlearninginductionheads}, generation maintaining coherency \cite{sia-duh-2023-context}, creation of task vectors based on in-context examples \cite{hendel-etal-2023-context}, \textit{etc.} The other stream focuses on the role of in-context examples, especially labels of these examples. \citeauthor{min-etal-2022-rethinking} \shortcite{min-etal-2022-rethinking} find that ground truth labels are not necessary and LLMs perform fairly well even with random labels. \citeauthor{wang-etal-2023-label} \shortcite{wang-etal-2023-label} find that label words play the role of anchors that aggregating information of the whole examples and serve as a reference for LLMs' final predictions. \citeauthor{wei2023largerlanguagemodelsincontext} \shortcite{wei2023largerlanguagemodelsincontext} find that larger language models can override semantic priors and learn from in-context examples with flipped labels or semantically-unrelated labels.

\section{Conclusion}

In this work, we analyze the effect of descriptive instructions in prompts during ICL and propose an \textit{Ensemble} prompt framework describing the properties of in-context examples selected by different methods. Experimental results on MT indicate that while LLMs are sensitive to prompt formats, they might not care the actual meaning of the descriptions and the framework improves LLMs' performance even with meaningless descriptions compared with the conventional prompt. We further apply the \textit{Ensemble} framework to four other NLP tasks and find that it achieves promising results, especially on small-scale models. These results suggest that rather than working hard on well-designed descriptions, making use of a proper prompt format would be more effective and efficient.

\section*{Acknowledgments}
This work is supported by the National Natural Science Foundation of China (62076008).

\section*{Limitations}
First, since there are so many open-source LLMs in the world nowadays, it is impossible to experiment with all existing models and thus our work only employ several commonly-used LLMs. Second, since we do not have access to the pre-training or post-training process of LLMs (either open-source or close-source), our analysis of the mechanism of ICL could be somewhat superficial. The behavior of LLMs can be highly subject to their training data, which we have no access to. Lastly, although we reveal that \textit{ERR} is a superior prompt format for several models, it could still be a local optimum and how to effectively search for a best prompt format for different models and tasks is still under-explored, which we leave for future work.

\bibliography{custom}

\appendix
\section{Prompts for MT Ablation Study}






\begin{figure}[htbp]
    \centering
    \includegraphics[width=1\linewidth]{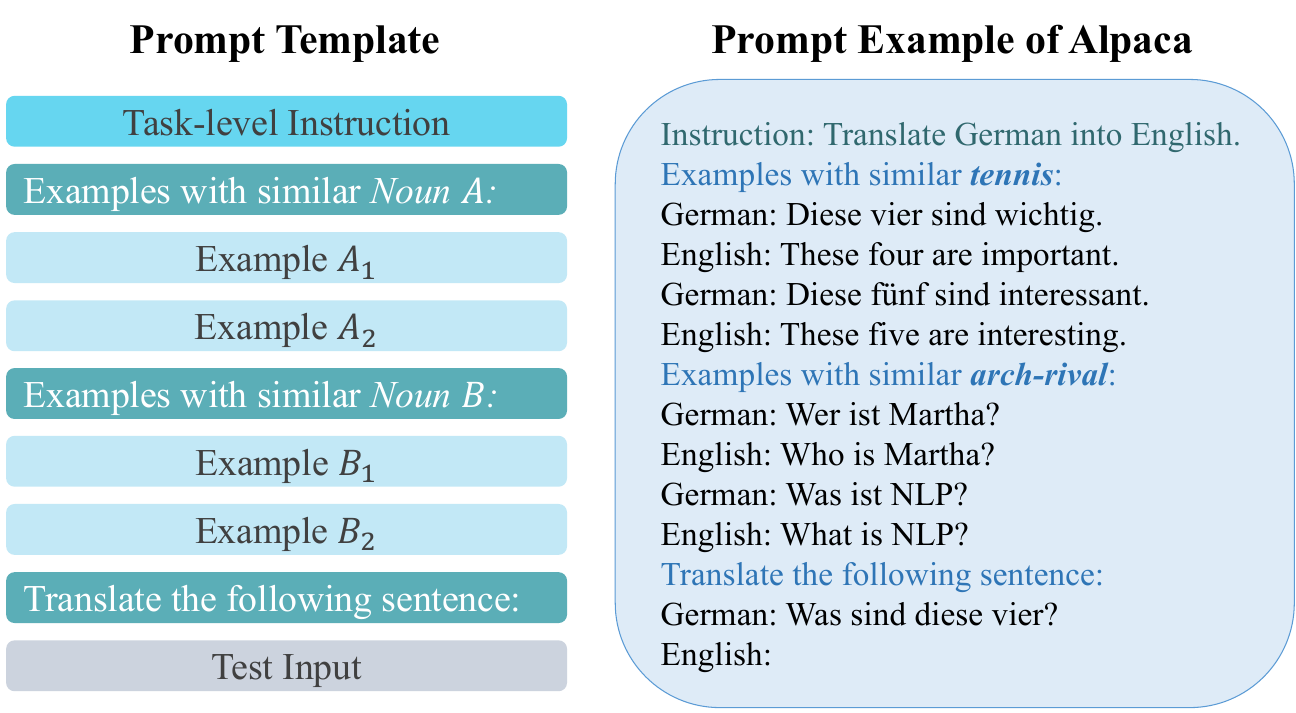}
    \caption{Template and example of \textit{Ensemble} (Random + Random).}
    \label{fig:tplt-ERR}
\end{figure}

\begin{figure}[htbp]
    \centering
    \includegraphics[width=1\linewidth]{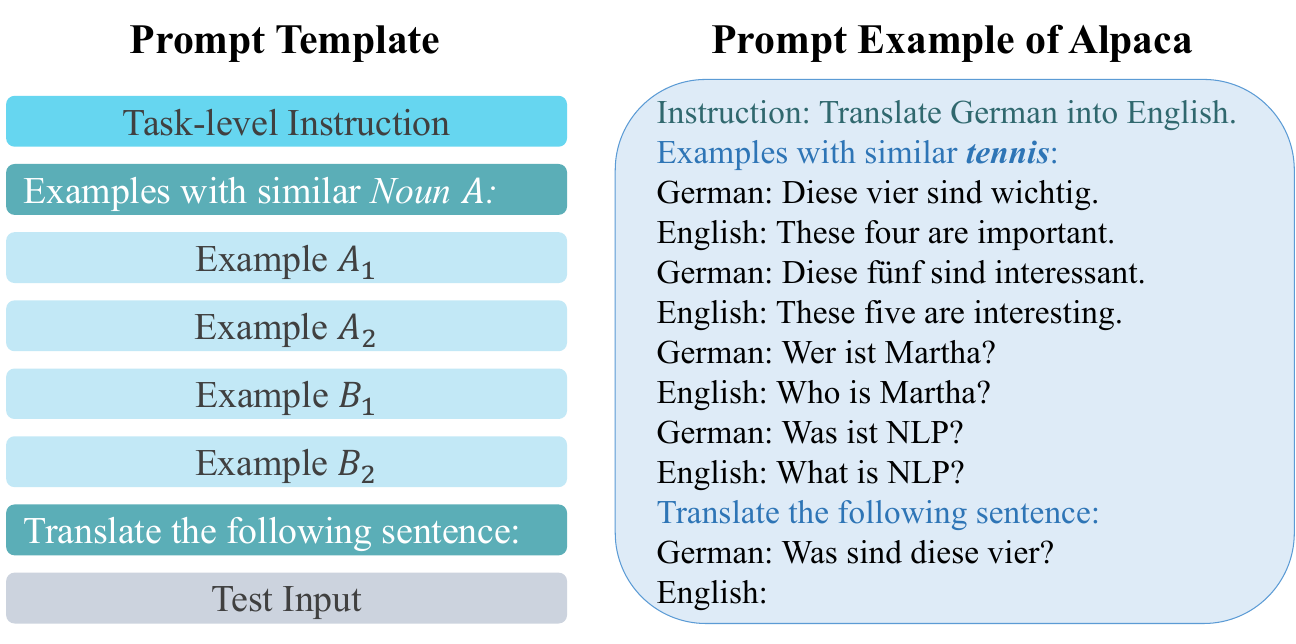}
    \caption{Template and example of \textit{Single} (Random).}
    \label{fig:tplt-SR}
\end{figure}

\begin{figure}[htbp]
    \centering
    \includegraphics[width=1\linewidth]{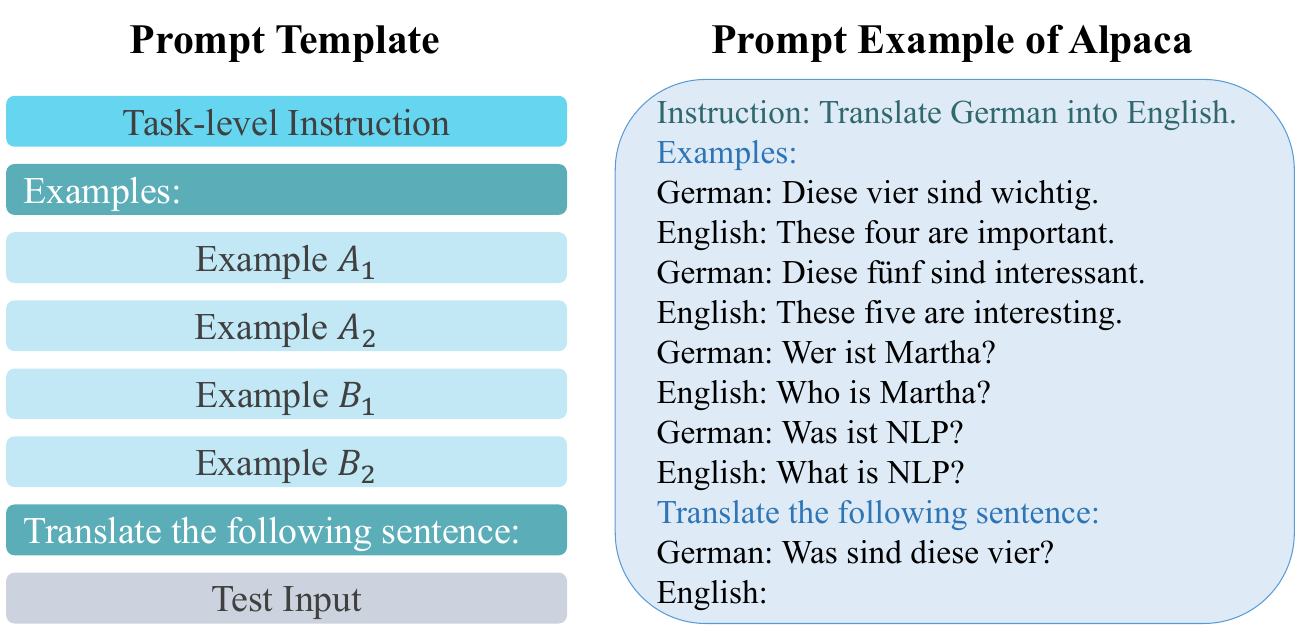}
    \caption{Template and example of \textit{Single} (Example).}
    \label{fig:tplt-SE}
\end{figure}

\begin{figure}[htbp]
    \centering
    \includegraphics[width=1\linewidth]{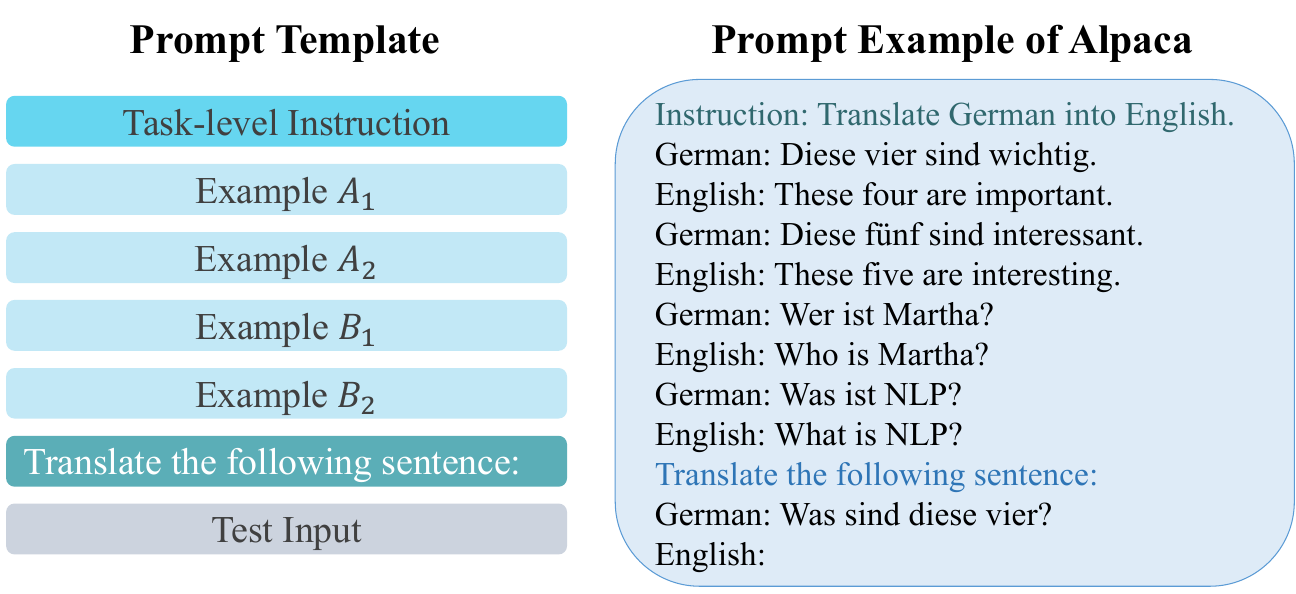}
    \caption{Template and example of \textit{Vanilla} (Translate).}
    \label{fig:tplt-VT}
\end{figure}


\label{subapp:prompt-ablation}
Templates and prompt examples of \textit{Ensemble} (Random + Random), \textit{Single} (Random), \textit{Single} (Example), \textit{Vanilla} (Translate) are shown in Figure \ref{fig:tplt-ERR}-\ref{fig:tplt-VT}.

\section{Full Experimental Results}
\label{sec:full}

\begin{table*}[htbp]
\small
\setlength{\tabcolsep}{1.5mm}
  \centering
    \begin{tabular}{c|c|c|ccc|ccc|c}
    \toprule
    \multirow{2}[2]{*}{\textbf{Prompt}} & \multirow{2}[2]{*}{\textbf{Selection-A}} & \multirow{2}[2]{*}{\textbf{Selection-B}} & \multicolumn{3}{c|}{\textbf{Into EN}} & \multicolumn{3}{c|}{\textbf{Out of EN}} & \multirow{2}[2]{*}{\textbf{Avg.}} \\
          &       &       & \textbf{DE} & \textbf{FR} & \textbf{RU} & \textbf{DE} & \textbf{FR} & \textbf{RU} &  \\
    \midrule
    \multirow{5}[2]{*}{\textit{Vanilla}} & Random & Random & 63.65 & 71.40  & 52.37 & 40.34 & 54.58 & 42.07 & 54.07 \\
          & BM25  & BM25  & 64.32 & 71.83 & 51.15 & 42.53 & 56.58 & 43.58 & 55.00 \\
          & Polynomial & Polynomial & 64.29 & 71.25 & 53.47 & 43.23 & 55.15 & 45.73 & 55.52 \\
          & BM25  & Polynomial & 65.34 & 72.01 & 53.74 & 43.41 & 56.48 & 46.04 & 56.17 \\
          & Polynomial & BM25  & 64.66 & 72.04 & 53.04 & 44.37 & 56.56 & 46.43 & 56.18 \\
    \midrule
    \multirow{5}[2]{*}{\textit{Ensemble} (Word + Syntax)} & Random & Random & 65.07 & 72.41 & 54.14 & 43.09 & 56.25 & 42.07 & 55.50 \\
          & BM25  & BM25  & 66.02 & 72.81 & 54.13 & 43.77 & 56.81 & 43.19 & 56.12 \\
          & Polynomial & Polynomial & 65.98 & 72.66 & 54.67 & 44.37 & 58.13 & 45.49 & 56.88 \\
          & BM25  & Polynomial & 66.08 & 72.58 & 54.29 & 43.51 & 57.07 & 45.26 & 56.47 \\
          & Polynomial & BM25  & 65.89 & 73.03 & 54.06 & 43.78 & 57.60 & 46.87 & 56.87 \\
    \midrule
    \multirow{5}[2]{*}{\textit{Ensemble} (Syntax + Word)} & Random & Random & 65.04 & 72.37 & 54.15 & 42.95 & 56.27 & 41.85 & 55.40 \\
          & BM25  & BM25  & 66.20 & 72.41 & 53.73 & 43.39 & 57.24 & 42.40 & 55.90 \\
          & Polynomial & Polynomial & 66.16 & 72.55 & 54.66 & 44.75 & 56.82 & 44.38 & 56.55 \\
          & BM25  & Polynomial & 65.99 & 72.77 & 54.21 & 44.05 & 57.26 & 45.09 & 56.56 \\
          & Polynomial & BM25  & 65.96 & 72.96 & 54.12 & 43.47 & 57.51 & 46.94 & 56.83 \\
    \midrule
    \multirow{5}[2]{*}{Diff. \textit{Ensemble} (Word + Syntax)} & Random & Random & 65.09 & 72.53 & 53.84 & 42.65 & 56.15 & 41.38 & 55.27 \\
          & BM25  & BM25  & 65.95 & 72.33 & 54.06 & 43.42 & 57.27 & 41.88 & 55.82 \\
          & Polynomial & Polynomial & 65.98 & 72.58 & 54.76 & 44.60 & 58.36 & 45.95 & 57.04 \\
          & BM25  & Polynomial & 66.04 & 72.49 & 54.75 & 44.03 & 57.70 & 45.95 & 56.83 \\
          & Polynomial & BM25  & 66.17 & 73.17 & 54.10 & 44.17 & 57.48 & 46.02 & 56.85 \\
    \midrule
    \multirow{5}[2]{*}{\textit{Ensemble} (Word + Semantics)} & Random & Random & 65.09 & 72.26 & 54.14 & 42.64 & 56.11 & 42.61 & 55.48 \\
          & BM25  & BM25  & 66.32 & 72.25 & 53.89 & 43.29 & 57.02 & 42.51 & 55.88 \\
          & Polynomial & Polynomial & 66.36 & 72.46 & 54.62 & 44.17 & 57.24 & 45.61 & 56.74 \\
          & BM25  & Polynomial & 65.80  & 72.74 & 54.40  & 44.06 & 56.86 & 45.54 & 56.57 \\
          & Polynomial & BM25  & 66.06 & 72.99 & 53.47 & 43.24 & 57.55 & 46.93 & 56.71 \\
    \midrule
    \multirow{5}[2]{*}{\textit{Ensemble} (Random + Random)} & Random & Random & 65.37 & 72.54 & 54.01 & 42.59 & 56.22 & 40.41 & 55.19 \\
          & BM25  & BM25  & 66.41 & 72.72 & 54.24 & 43.08 & 56.59 & 41.44 & 55.75 \\
          & Polynomial & Polynomial & 66.28 & 72.56 & 54.70  & 43.81 & 57.42 & 43.22 & 56.33 \\
          & BM25  & Polynomial & 65.99 & 72.77 & 54.04 & 44.85 & 57.25 & 45.39 & 56.72 \\
          & Polynomial & BM25  & 65.73 & 73.26 & 53.73 & 43.27 & 56.78 & 44.73 & 56.25 \\
    \bottomrule
    \end{tabular}%
  \caption{Full MT results of XGLM.}
  \label{tab:full-xglm}%
\end{table*}%

\begin{table*}[htbp]
\small
\setlength{\tabcolsep}{1.5mm}
  \centering
    \begin{tabular}{c|c|c|ccc|ccc|c}
    \toprule
    \multirow{2}[2]{*}{\textbf{Prompt}} & \multirow{2}[2]{*}{\textbf{Selection-A}} & \multirow{2}[2]{*}{\textbf{Selection-B}} & \multicolumn{3}{c|}{\textbf{Into EN}} & \multicolumn{3}{c|}{\textbf{Out of EN}} & \multirow{2}[2]{*}{\textbf{Avg.}} \\
          &       &       & \textbf{DE} & \textbf{FR} & \textbf{RU} & \textbf{DE} & \textbf{FR} & \textbf{RU} &  \\
    \midrule
    \multirow{5}[2]{*}{\textit{Vanilla}} & Random & Random & 69.88 & 76.46 & 57.80 & 42.52 & 56.61 & 29.25 & 55.42 \\
          & BM25  & BM25  & 69.08 & 76.41 & 58.52 & 43.65 & 57.34 & 32.63 & 56.27 \\
          & Polynomial & Polynomial & 69.65 & 75.79 & 58.77 & 43.55 & 56.60 & 32.39 & 56.13 \\
          & BM25  & Polynomial & 69.34 & 76.02 & 58.38 & 43.31 & 56.79 & 33.21 & 56.18 \\
          & Polynomial & BM25  & 69.03 & 76.11 & 57.84 & 42.20 & 55.81 & 31.93 & 55.49 \\
    \midrule
    \multirow{5}[2]{*}{\textit{Ensemble} (Word + Syntax)} & Random & Random & 69.86 & 76.64 & 57.57 & 43.51 & 57.25 & 30.79 & 55.94 \\
          & BM25  & BM25  & 69.44 & 76.21 & 57.60 & 44.41 & 58.16 & 31.26 & 56.18 \\
          & Polynomial & Polynomial & 69.86 & 76.06 & 58.26 & 44.46 & 57.06 & 33.27 & 56.50 \\
          & BM25  & Polynomial & 69.58 & 76.09 & 57.98 & 43.84 & 57.57 & 32.85 & 56.32 \\
          & Polynomial & BM25  & 69.41 & 76.32 & 57.78 & 42.79 & 58.18 & 31.07 & 55.93 \\
    \midrule
    \multirow{5}[2]{*}{\textit{Ensemble} (Syntax + Word)} & Random & Random & 69.80 & 76.73 & 57.53 & 43.50 & 57.31 & 30.70 & 55.93 \\
          & BM25  & BM25  & 69.46 & 76.15 & 57.63 & 44.58 & 58.46 & 32.09 & 56.40 \\
          & Polynomial & Polynomial & 69.76 & 76.11 & 58.14 & 44.11 & 56.64 & 33.24 & 56.33 \\
          & BM25  & Polynomial & 69.60 & 76.11 & 57.91 & 43.88 & 57.81 & 32.22 & 56.26 \\
          & Polynomial & BM25  & 69.64 & 76.11 & 57.77 & 42.42 & 58.43 & 31.41 & 55.96 \\
    \midrule
    \multirow{5}[2]{*}{Diff. \textit{Ensemble} (Word + Syntax)} & Random & Random & 69.77 & 76.63 & 57.46 & 43.67 & 57.48 & 30.55 & 55.93 \\
          & BM25  & BM25  & 69.44 & 76.33 & 57.70 & 44.24 & 58.48 & 31.76 & 56.33 \\
          & Polynomial & Polynomial & 69.75 & 76.07 & 58.12 & 44.09 & 57.31 & 32.46 & 56.30 \\
          & BM25  & Polynomial & 69.54 & 76.21 & 57.68 & 43.61 & 57.49 & 32.27 & 56.13 \\
          & Polynomial & BM25  & 69.49 & 76.23 & 57.65 & 42.79 & 58.20 & 32.09 & 56.08 \\
    \midrule
    \multirow{5}[2]{*}{\textit{Ensemble} (Word + Semantics)} & Random & Random & 69.88 & 76.70 & 57.51 & 43.30 & 57.35 & 30.98 & 55.96 \\
          & BM25  & BM25  & 69.44 & 76.20 & 57.65 & 44.47 & 57.88 & 32.48 & 56.35 \\
          & Polynomial & Polynomial & 69.82 & 76.10 & 58.35 & 44.00 & 57.25 & 33.45 & 56.50 \\
          & BM25  & Polynomial & 69.57 & 76.23 & 58.12 & 44.07 & 57.73 & 32.79 & 56.42 \\
          & Polynomial & BM25  & 69.49 & 76.17 & 57.99 & 43.20 & 58.49 & 31.66 & 56.17 \\
    \midrule
    \multirow{5}[2]{*}{\textit{Ensemble} (Random + Random)} & Random & Random & 69.77 & 76.61 & 57.38 & 43.52 & 57.55 & 30.90 & 55.95 \\
          & BM25  & BM25  & 69.46 & 76.24 & 57.58 & 44.40 & 58.25 & 33.23 & 56.53 \\
          & Polynomial & Polynomial & 69.63 & 75.93 & 57.77 & 44.36 & 56.73 & 33.45 & 56.31 \\
          & BM25  & Polynomial & 69.55 & 76.01 & 57.86 & 42.75 & 57.80 & 32.63 & 56.10 \\
          & Polynomial & BM25  & 69.74 & 76.04 & 57.73 & 43.33 & 58.30 & 33.27 & 56.40 \\
    \bottomrule
    \end{tabular}%
  \caption{Full MT results of Alpaca.}
  \label{tab:full-alpaca}%
\end{table*}%

\begin{table*}[htbp]
\small
\setlength{\tabcolsep}{1mm}
  \centering
    \begin{tabular}{c|c|cccccccccc}
    \toprule
    \multirow{2}[2]{*}{\textbf{Model}} & \multicolumn{1}{c|}{\multirow{2}[2]{*}{\textbf{Template}}} & \multicolumn{10}{c}{\textbf{Performance}} \\
          &       & \textbf{Date} & \textbf{SgyQA} & \textbf{CSQA} & \textbf{Sports} & \textbf{LF} & \textbf{TO} & \textbf{GSM8K} & \textbf{AQuA} & \textbf{KU} & \textbf{Avg.} \\
    \midrule
    \multirow{4}[2]{*}{LLaMA-2-7B} & \textit{Vanilla} w/ CoT & 3.01  & 50.99  & 21.70  & 50.20  & 48.72  & 30.74  & \textbf{22.29 } & 21.65  & 45.24  & 32.73  \\
          & \textit{ERR} w/ CoT & \textbf{45.21 } & 59.78  & \textbf{55.36 } & \textbf{79.42 } & \textbf{56.03 } & \textbf{41.55 } & 21.15  & \textbf{23.62 } & \textbf{85.71 } & \textbf{51.98 } \\
          & \textit{Vanilla} w/o CoT & 0.27  & 47.33  & 19.25  & 54.42  & 48.91  & 30.07  & 0.91  & 23.23  & 52.38  & 30.75  \\
          & \textit{ERR} w/o CoT & 27.67  & \textbf{59.88 } & 23.91  & 16.77  & 1.58  & 36.49  & 5.91  & 22.44  & 69.05  & 29.30  \\
    \midrule
    \multirow{4}[2]{*}{LLaMA-3.1-8B} & \textit{Vanilla} w/ CoT & 9.04  & 51.28  & 24.41  & 55.32  & \textbf{52.08 } & 35.47  & \textbf{49.05 } & 24.80  & \textbf{69.05 } & 41.17  \\
          & \textit{ERR} w/ CoT & \textbf{38.63 } & 55.53  & 31.04  & \textbf{74.60 } & 51.58  & \textbf{38.18 } & 48.52  & \textbf{26.77 } & \textbf{69.05 } & \textbf{48.21 } \\
          & \textit{Vanilla} w/o CoT & 7.95  & 41.90  & 28.42  & 54.22  & 51.98  & 37.50  & 1.14  & 16.93  & 28.57  & 29.84  \\
          & \textit{ERR} w/o CoT & 13.70  & \textbf{59.68 } & \textbf{36.12 } & 50.00  & 50.79  & 35.81  & 4.70  & 22.83  & 57.14  & 36.75  \\
    \midrule
    \multirow{4}[2]{*}{Alpaca-7B} & \textit{Vanilla} w/ CoT & 26.03  & 60.38  & \textbf{50.53 } & \textbf{80.12 } & 52.27  & \textbf{35.81 } & 6.44  & 18.11  & \textbf{71.43 } & \textbf{44.57 } \\
          & \textit{ERR} w/ CoT & \textbf{26.85 } & 60.47  & 50.04  & 74.40  & 54.74  & 35.14  & \textbf{6.75 } & 20.08  & \textbf{71.43 } & 44.43  \\
          & \textit{Vanilla} w/o CoT & 25.21  & 60.77  & 46.27  & 60.24  & \textbf{55.63 } & 35.47  & 4.70  & \textbf{24.80 } & 64.29  & 41.93  \\
          & \textit{ERR} w/o CoT & 24.93  & \textbf{61.17 } & 50.12  & 61.04  & 55.14  & 34.80  & 4.78  & 24.41  & 64.29  & 42.30  \\
    \midrule
    \multirow{4}[2]{*}{Mistral-7B} & \textit{Vanilla} w/ CoT & 6.30  & 57.71  & 50.37  & 61.24  & 48.22  & 31.08  & 43.21  & \textbf{29.13 } & 73.81  & 44.56  \\
          & \textit{ERR} w/ CoT & \textbf{36.99 } & \textbf{66.60 } & \textbf{61.83 } & \textbf{84.94 } & \textbf{64.03 } & 33.78  & \textbf{43.97 } & 25.98  & \textbf{83.33 } & \textbf{55.72 } \\
          & \textit{Vanilla} w/o CoT & 0.27  & 49.70  & 20.97  & 78.21  & 46.05  & 31.08  & 1.06  & 18.90  & 61.90  & 34.24  \\
          & \textit{ERR} w/o CoT & 13.15  & 64.13  & 55.86  & 77.11  & 61.86  & \textbf{49.32 } & 7.96  & 22.44  & 66.67  & 46.50  \\
    \bottomrule
    \end{tabular}%
  \caption{Full results of Date Understanding (Date), StrategyQA (SgyQA), CSQA, Sports Understanding (Sports), Logical Fallacy (LF), Three Objects (TO), GSM8K, AQuA and Known Unknowns (KU).}
  \label{tab:full-reasoning}%
\end{table*}%

Full results of MT are presented in Table~\ref{tab:full-xglm} and~\ref{tab:full-alpaca}. Full results of other tasks in Section~\ref{sec:small-reasoning} are presented in Table~\ref{tab:full-reasoning}.

\section{Attention Weights on Alpaca}
\label{sec:attn-alpaca}

\begin{figure*}
    \centering
    \includegraphics[width=1\linewidth]{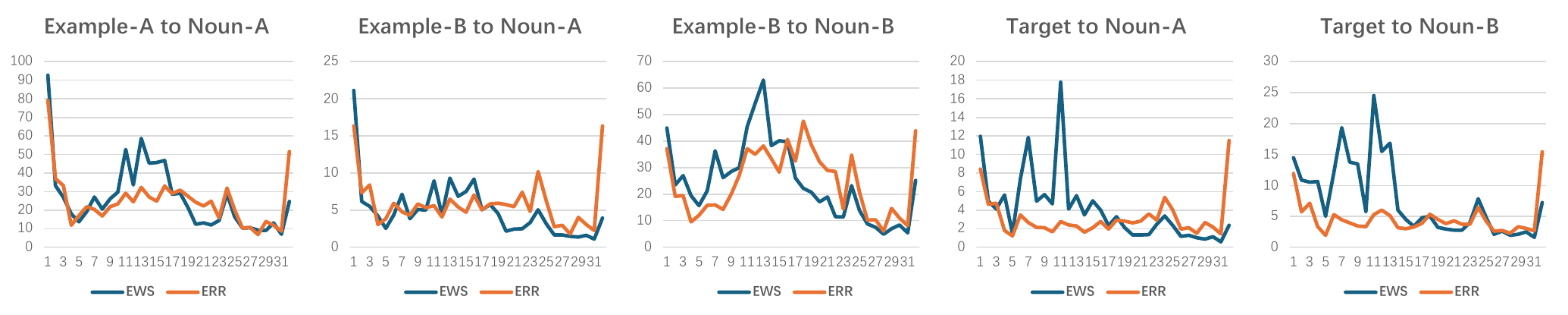}
    \caption{Attention weights (×1e-4) on Alpaca of all 32 layers with BM25 + Polynomial examples. EWS and \textit{ERR} denotes \textit{Ensemble} (Word + Syntax) and \textit{Ensemble} (Random + Random) respectively.}
    \label{fig:attn-alpaca}
\end{figure*}

Figure \ref{fig:attn-alpaca} presents the attention weights on Alpaca. For example-to-noun attention weights, \textit{ERR} is close to EWS. For target-to-noun attention weights, EWS is higher in shallow layers but falls behind \textit{ERR} in deeper layers, especially in the last layer. This demonstrates that Alpaca might pay more attention to the meaningful words ("word" and "syntax") when understanding the context in shallow layers but gradually forgets them when it comes to generation in the deeper layers. In short, EWS performs no higher than \textit{ERR} in most cases.

\section{Dataset Details for Reasoning Tasks}
\label{app:data}
We list the details of splitting training set (example database) and test set for our conducted reasoning tasks, covering four types and nine datasets. We set random seed for all possible shuffling and sampling operations to $42$. Note that we experiment with \textbf{4-shot} for all datasets.

\subsection{Datasets Fetched from Exclusive Source}
\begin{itemize}
    \item CSQA \cite{csqa}: \url{https://www.tau-nlp.org/commonsenseqa}. We follow the official split and select the training set as our example database and the dev set as our test set. Because the training set itself is randomly divided from the whole dataset, we directly select examples from it in order.

    \item GSM8K \cite{gsm8k}: \url{https://github.com/openai/grade-school-math}. We select the \texttt{test.jsonl} as our test set and the \texttt{train.jsonl} as our example database and randomly sample four examples from it.

    \item AQuA \cite{aqua}: \url{https://github.com/google-deepmind/AQuA}. We select the \texttt{test.json} as our test set. Since the original training set is relatively large, for simplicity, we directly copy the four examples listed in the supplementary materials of \citet{cot} and we ensure that these four examples do not appear in the test set.

\end{itemize}

\subsection{Datasets Fetched from The Big-bench}
For StrategyQA \cite{strategyqa}, Date, Sports, Logical Fallacy, Three Objects, and Known Unknowns, we fetched them from the Big-bench \cite{bigbench}. Each of them has a \texttt{task.json}. We randomly shuffle the \texttt{task.json} and split it to a training set and a test set. Then we select examples from the training set in order. 

Specifically, the principle for splitting the training and test sets is as follows: If the total number of samples exceeds 1,012 a lot, we retain 1,012 samples as the test set and use the remainder as the training set. Otherwise, we select four examples for the training set and use the rest for testing. For the Sports and Logical Fallacy datasets, which have only two possible answers (similar to binary classification), we first separate the positive and negative examples, shuffle them individually, and then construct the test set and training set. The test set is composed of an equal number of positive and negative examples, with the remaining samples used as the training set.


\section{Prompts for Reasoning Tasks Used in this Work}

\label{app:prompt-reasoning}

\begin{figure*}[htbp]
    \centering
    \includegraphics[width=1\linewidth]{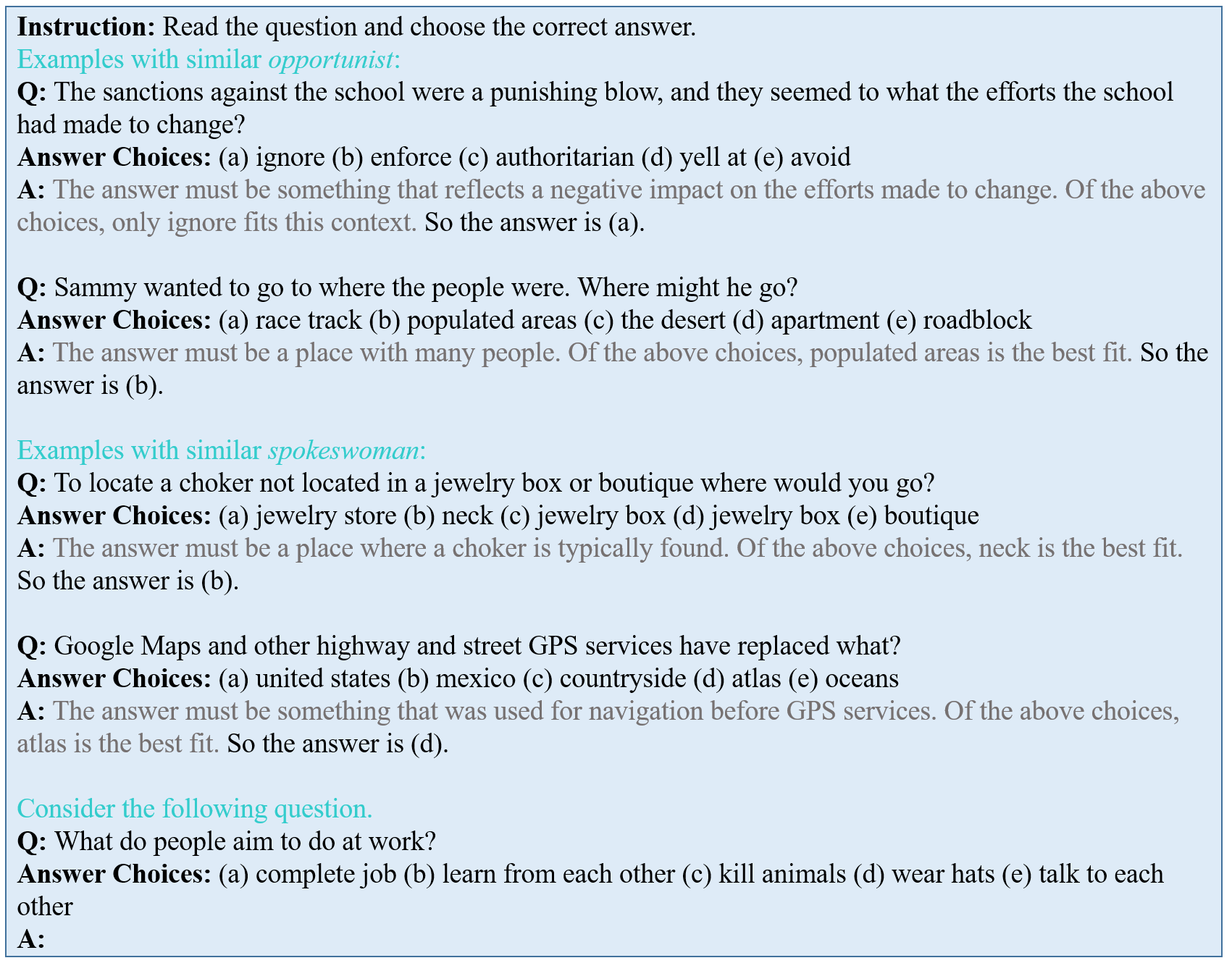}
    \caption{Prompt for CSQA.}
    \label{fig:csqa}
\end{figure*}

\begin{figure*}[htbp]
    \centering
    \includegraphics[width=1\linewidth]{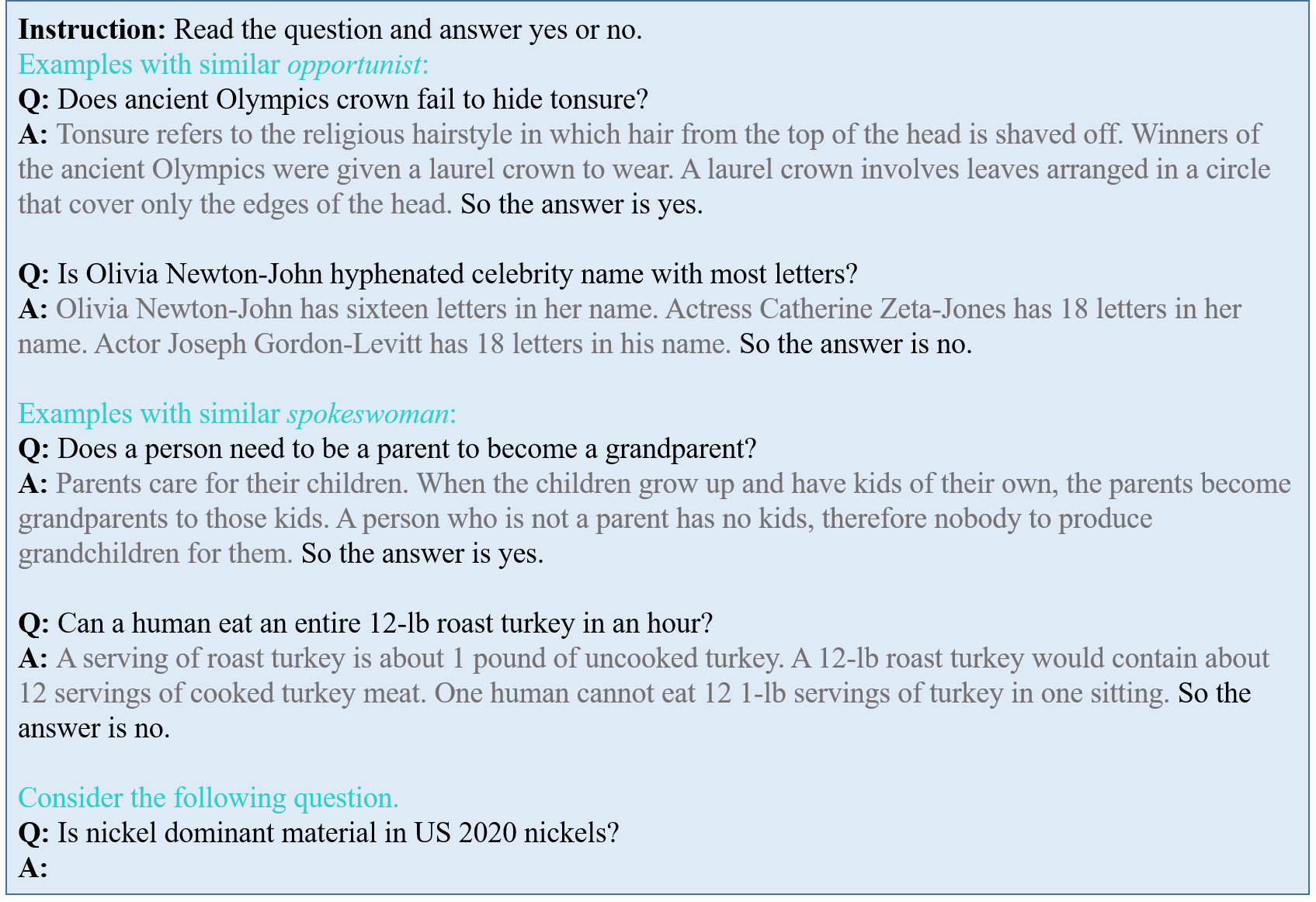}
    \caption{Prompt for StrategyQA.}
    \label{fig:strategyqa}
\end{figure*}

\begin{figure*}[htbp]
    \centering
    \includegraphics[width=1\linewidth]{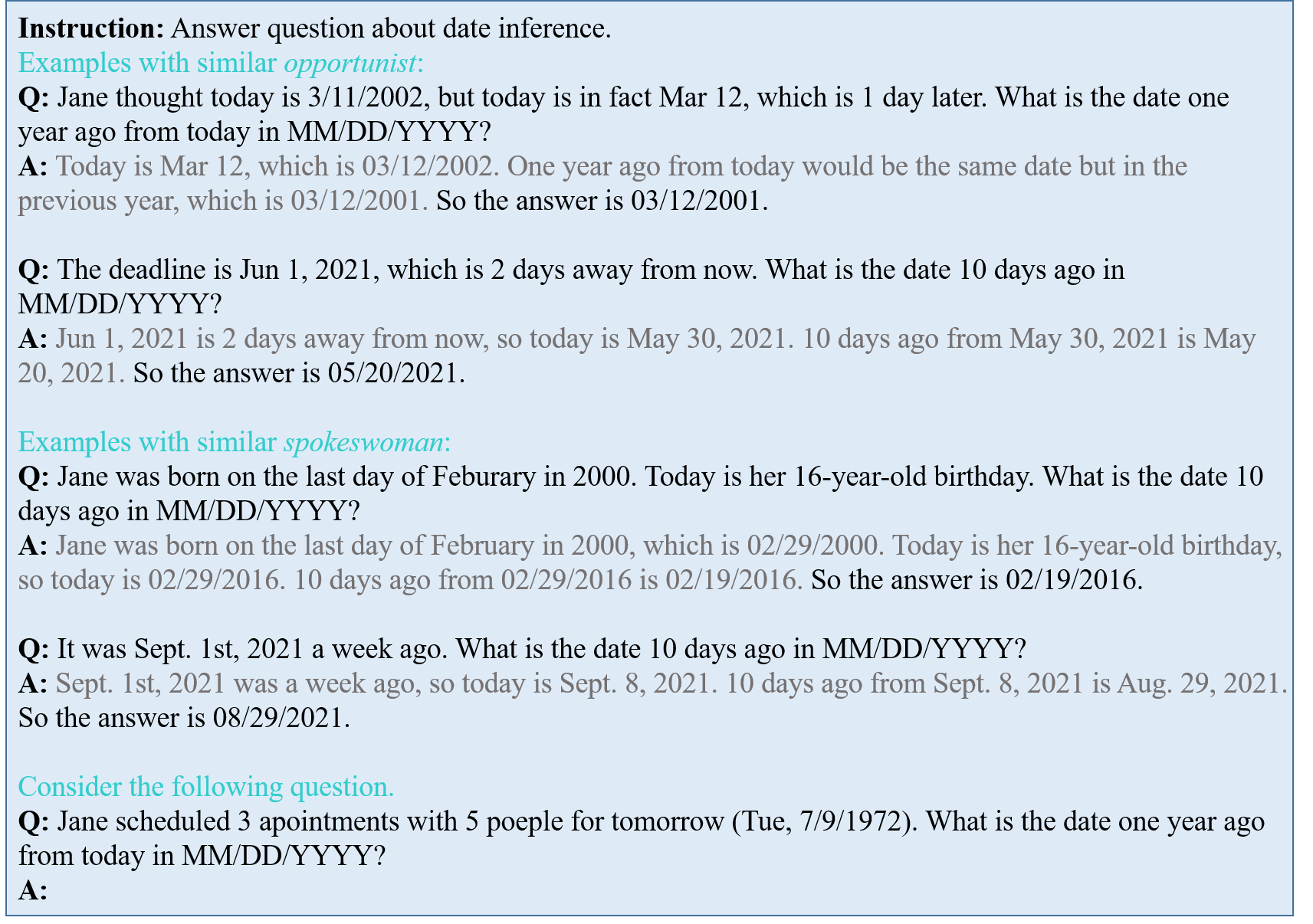}
    \caption{Prompt for Date.}
    \label{fig:date}
\end{figure*}

\begin{figure*}[htbp]
    \centering
    \includegraphics[width=1\linewidth]{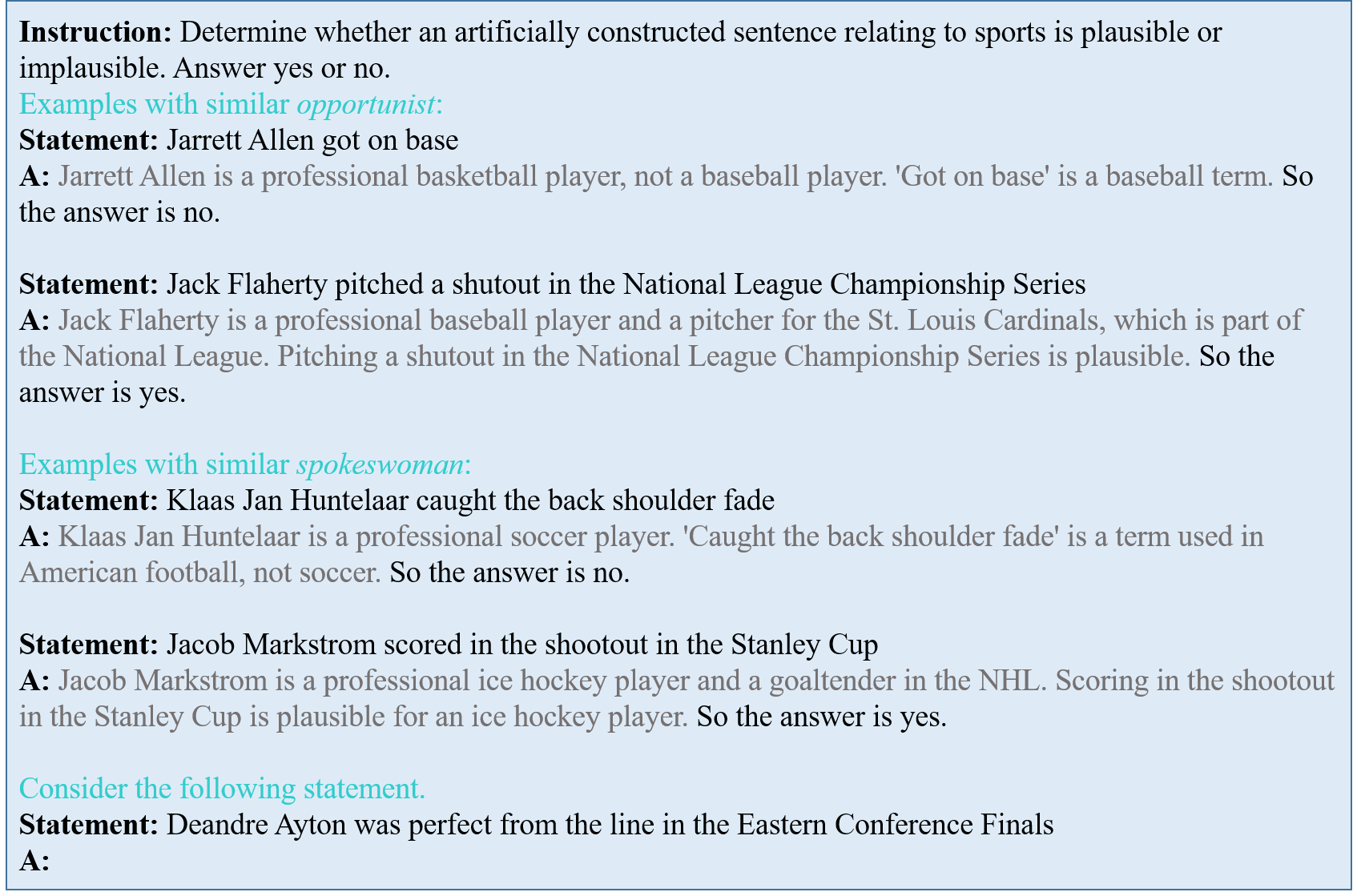}
    \caption{Prompt for Sports.}
    \label{fig:sports}
\end{figure*}

\begin{figure*}[htbp]
    \centering
    \includegraphics[width=1\linewidth]{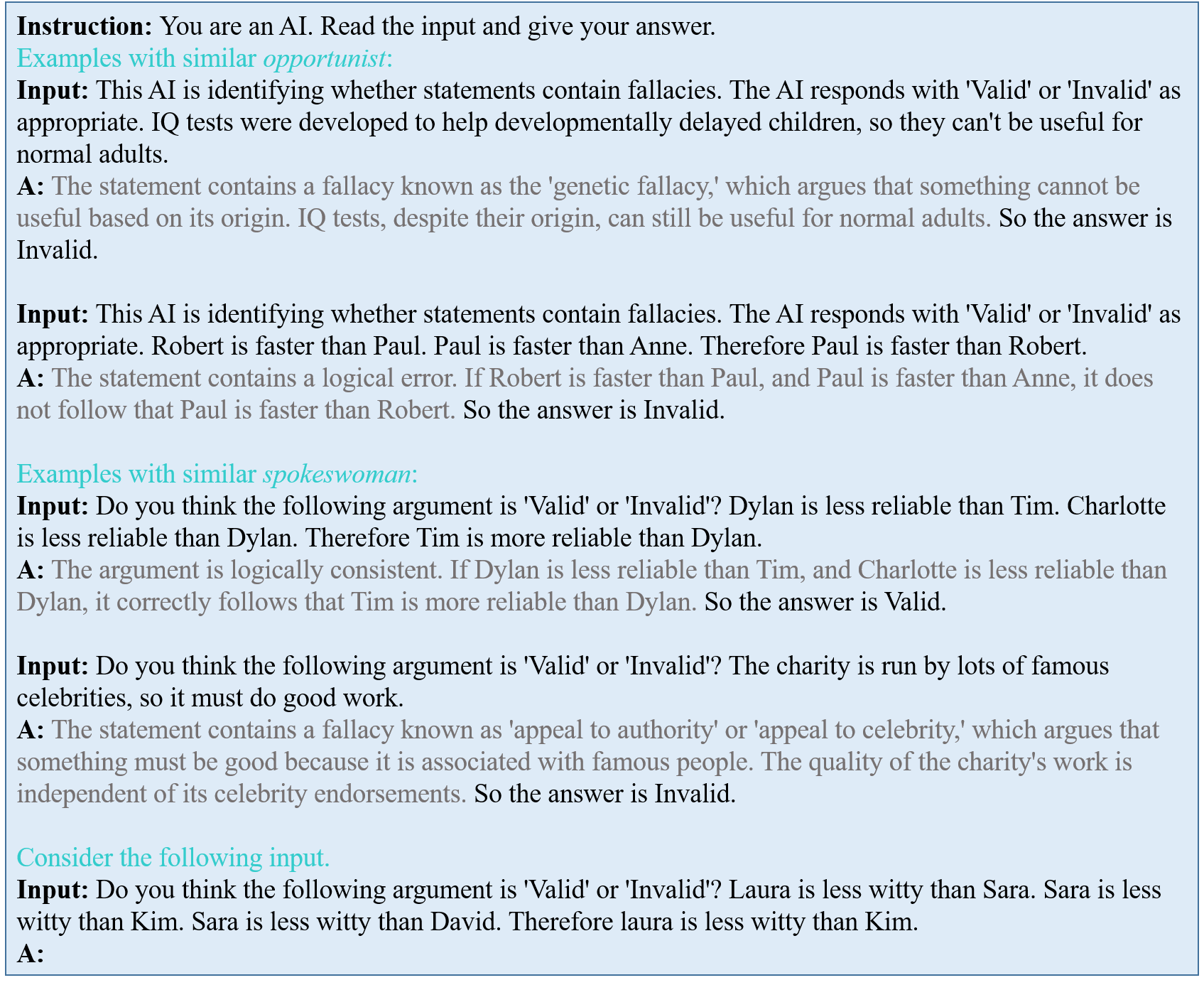}
    \caption{Prompt for Logical Fallacy.}
    \label{fig:logical_fallacy}
\end{figure*}

\begin{figure*}[htbp]
    \centering
    \includegraphics[width=1\linewidth]{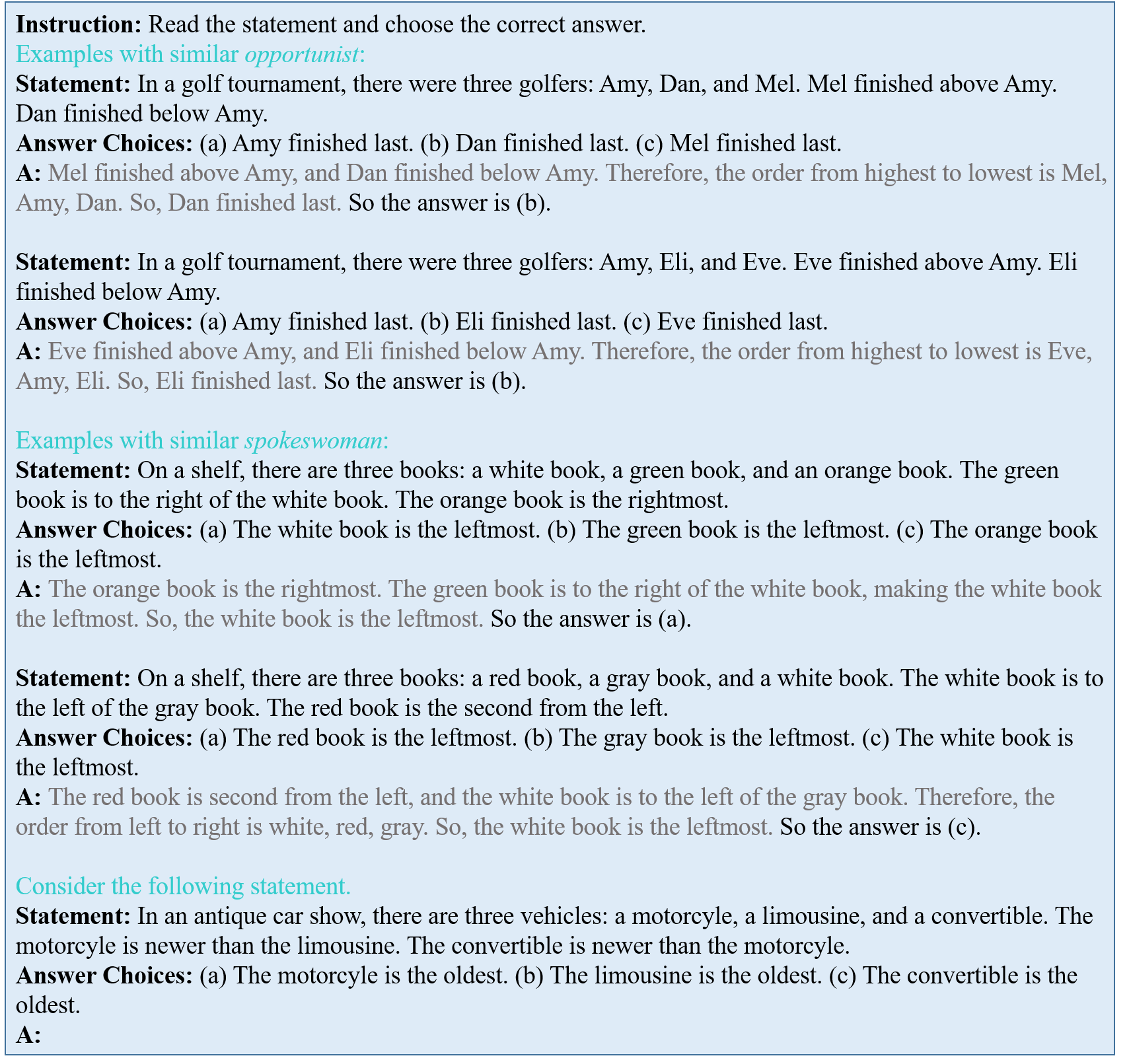}
    \caption{Prompt for Three Objects.}
    \label{fig:three_objects}
\end{figure*}

\begin{figure*}[htbp]
    \centering
    \includegraphics[width=1\linewidth]{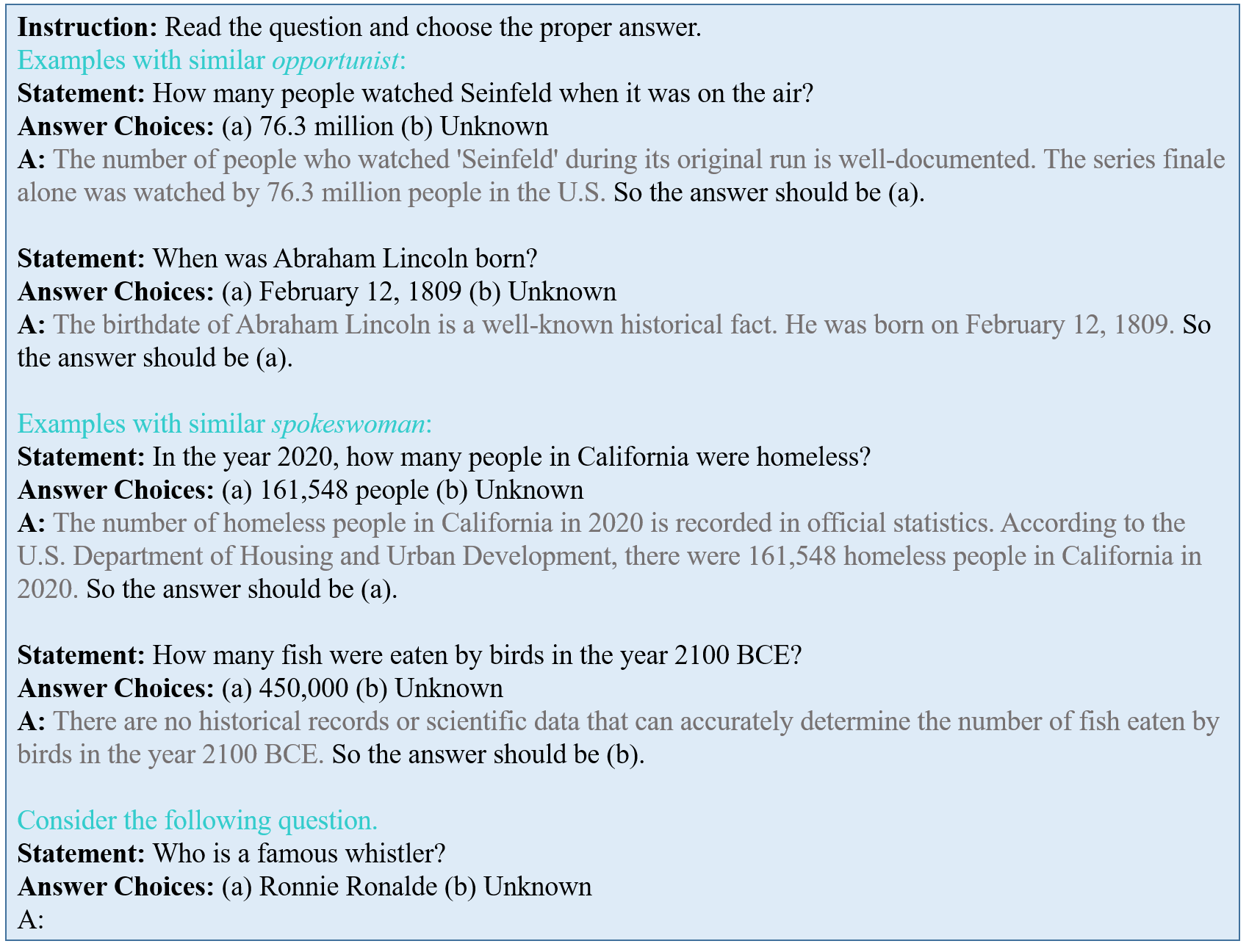}
    \caption{Prompt for Known Unknowns.}
    \label{fig:hallucination}
\end{figure*}

\begin{figure*}[htbp]
    \centering
    \includegraphics[width=1\linewidth]{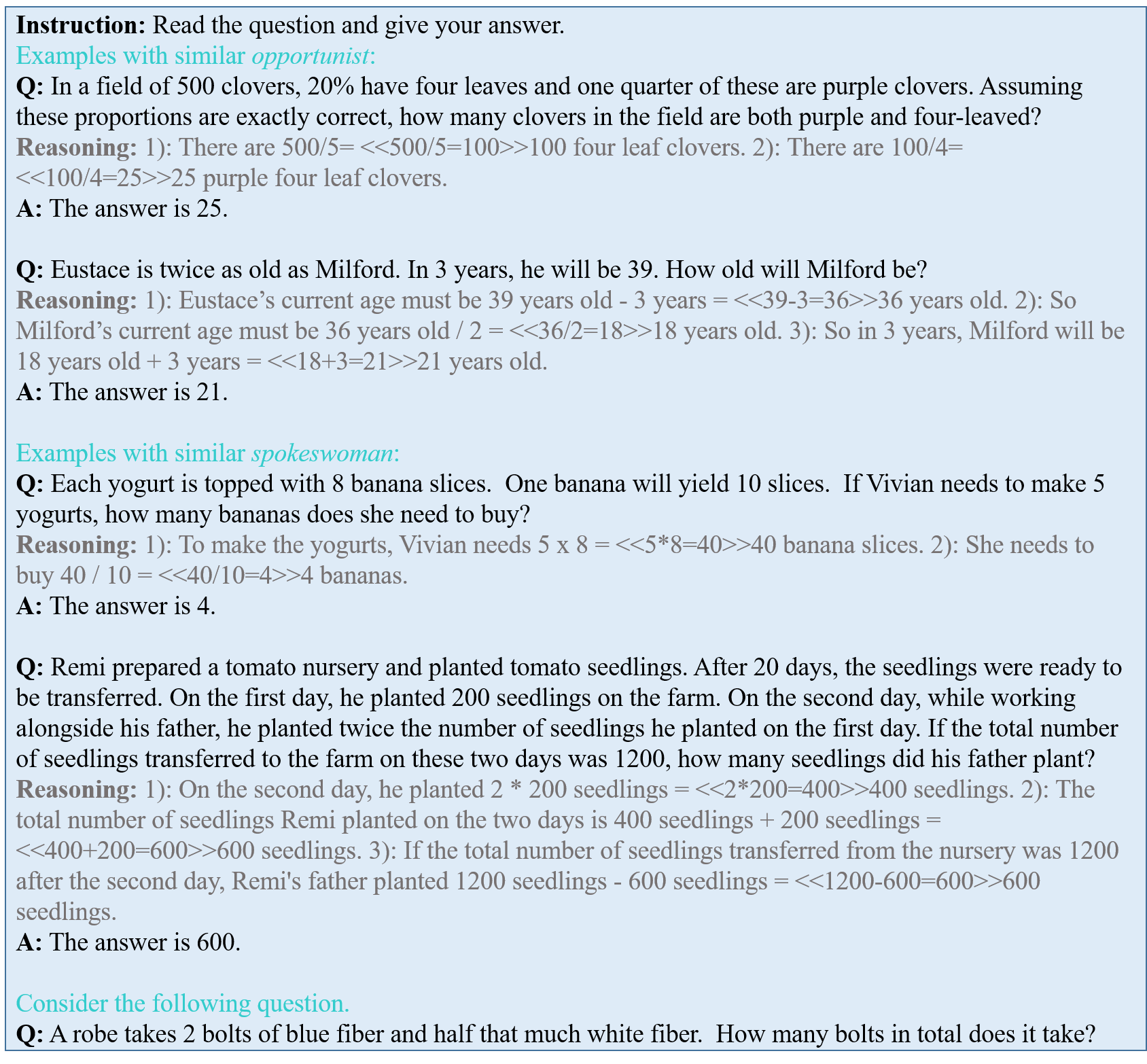}
    \caption{Prompt for GSM8K. For this dataset, we let LLMs first generate reasoning and then answer under the "w/ CoT" setting.}
    \label{fig:gsm8k}
\end{figure*}

\begin{figure*}[htbp]
    \centering
    \includegraphics[width=1\linewidth]{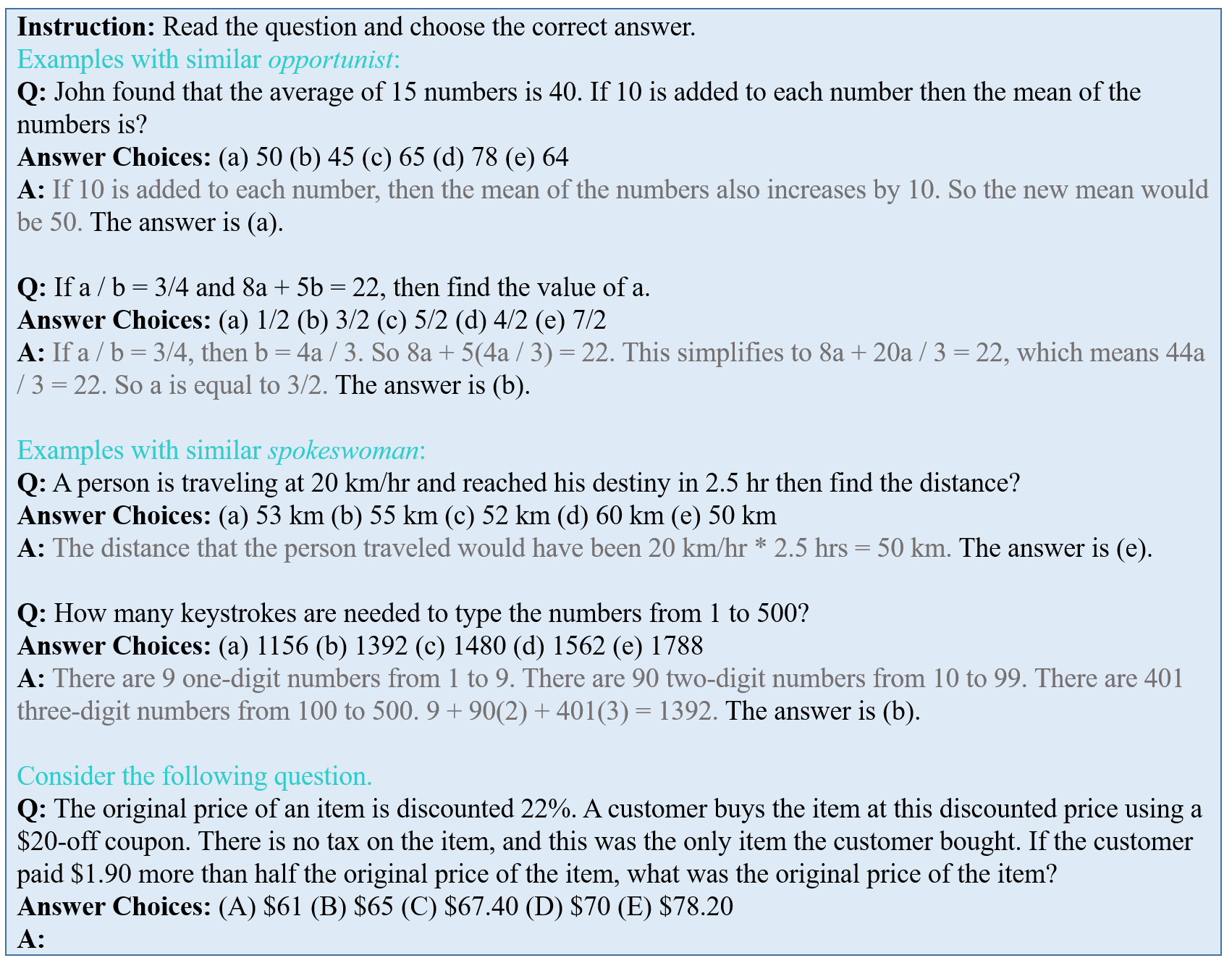}
    \caption{Prompt for AQuA.}
    \label{fig:aqua}
\end{figure*}

Figure \ref{fig:csqa}-\ref{fig:aqua} show the examples of \textit{ERR} (w/ CoT) prompt for respective datasets. Some tasks contain \textbf{Answer Choices}. In order to save space, the blank lines between the options are replaced with spaces in those figures. Each Figure has grey text for reasoning, cyan text for the parts of \textit{ERR} that are unique to \textit{Vanilla}, and italic words in the cyan text representing random nouns. Therefore, deleting the grey text gives \textit{ERR} (w/o CoT), keeping the grey text but deleting the cyan text gives \textit{Vanilla} (w/ CoT), and deleting both the cyan and grey text gives \textit{Vanilla} (w/o CoT).\footnote{When changing "w/ CoT" to "w/o CoT", you may also need to replace "So the answer is ..." with "The answer is ..." for syntactical reasons.} The reasoning is generated by ChatGPT \footnote{https://chatgpt.com/}. Note the ChatGPT is not the same as GPT-3.5 we used for experiments.

\section{Results of Llama2}
\label{app:llama2}

\begin{figure*}
    \centering
    \includegraphics[width=1.0\linewidth]{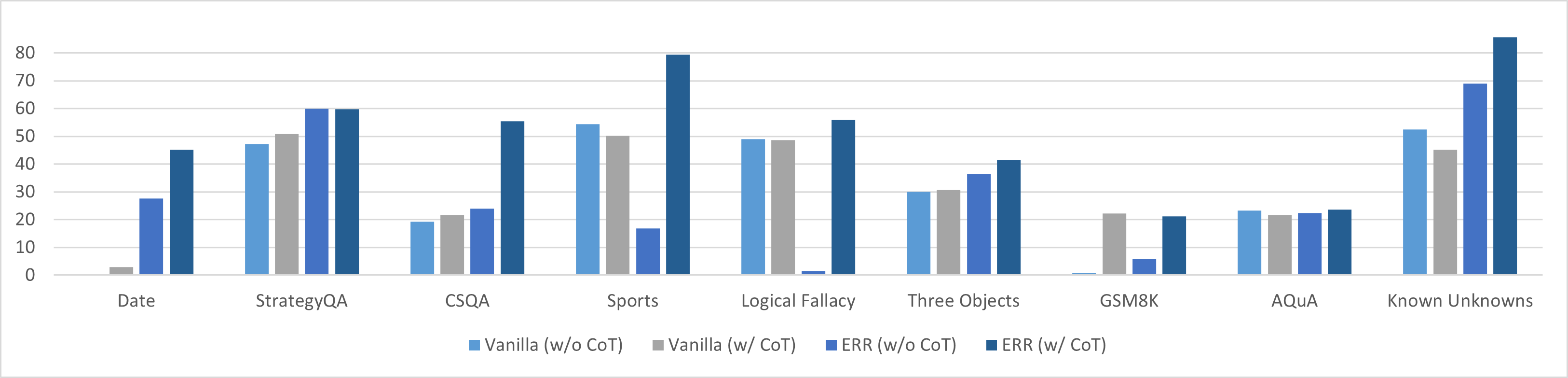}
    \caption{Results of Llama2 on the nine datasets.}
    \label{fig:llama2}
\end{figure*}

Results of Llama2-7B-chat-hf \cite{llama2} on the nine datasets are presented in Figure \ref{fig:llama2}. While \textit{ERR} outperforms \textit{Vanilla} with Llama2 across most datasets, its performance on Logical Fallacy and Sports is notably poor. Llama2 almost always responds with confused emojis for Logical Fallacy questions and outputs questions like "plausible or implausible?" for Sports, leading to predominantly incorrect answers. Further investigation into these issues is left for future work.

\section{Computational Details}
\subsection{Hardware}
Inference of LLMs runs on an NVIDIA A40 GPU (with memory of 48 GB). Other experiments run on Intel$^\circledR$ Xeon$^\circledR$ Gold 6348 CPU (with memory of 256 GB).

\subsection{Software}
Our OS: Ubuntu 20.04.6 LTS. Our code: Python only. Libraries and packages are specified in the source code.

\section{Licenses}
\begin{table}[htbp]
\small
  \centering
    \begin{tabular}{ll}
    \toprule
    \textbf{Artifact} & \textbf{License} \\
    \midrule
    XGLM  & MIT \\
    Alpaca & Apache-2.0 \\
    Llama & Llama Community License Agreement\\
    Mistral & Apache-2.0\\
    COMET & Apache-2.0 \\
    FLORES-101 & CC-BY-SA-4.0 \\
    Europarl & Unknown \\
    ParaCrawl & CC0 \\
    CSQA&CC-BY-SA-4.0\\
StrategyQA&MIT\\
BIG-bench&Apache-2.0\\
GSM8K&MIT\\
AQuA&Apache-2.0\\

    \bottomrule
    \end{tabular}%
  \caption{Licenses of scientific artifacts we use.}
  \label{tab:artifact}%
\end{table}%

The licenses of the scientific artifacts we use are shown in Table \ref{tab:artifact}.

\end{document}